\documentclass[10pt,twocolumn,letterpaper]{article}

\usepackage{cvpr}              
\usepackage{multirow}
\usepackage{makecell}
\usepackage{booktabs}
\usepackage{overpic}
\usepackage{adjustbox}
\usepackage{caption}
\usepackage{color}
\usepackage{pifont}

\definecolor{cvprblue}{rgb}{0.21,0.49,0.74}
\usepackage[pagebackref,breaklinks,colorlinks,allcolors=cvprblue]{hyperref}

\hypersetup{
    colorlinks=true,
    urlcolor=magenta,
    linkcolor=blue,
    citecolor=blue
}

\def\confName{CVPR}
\def\confYear{2026}

\title{CASR: A Robust Cyclic Framework for Arbitrary Large-Scale Super-Resolution with Distribution Alignment and Self-Similarity Awareness
}

\begin{document}

\author{
    Wenhao Guo$^{1}$, Zhaoran Zhao$^{1}$, Peng Lu$^{1,*}$,  Sheng Li$^{2}$, Qian Qiao$^{1}$, DeRui Li$^{1}$\\[4pt]
    \small $^{1}$Beijing University of Posts and Telecommunications \quad $^{2}$Peking University \\[2pt]
    \small \texttt{\{whguo, zhaozhaoran, lupeng, qqiao, deruili\}@bupt.edu.cn, lisheng@pku.edu.cn}
}

\maketitle
\renewcommand*{\thefootnote}{}
\footnote{$^*$ Corresponding author.}

\begin{abstract}
Arbitrary-Scale SR (ASISR) remains fundamentally limited by cross-scale distribution shift: once the inference scale leaves the training range, noise, blur, and artifacts accumulate sharply. We revisit this challenge from a cross-scale distribution transition perspective and propose CASR, a simple yet highly efficient cyclic SR framework that reformulates ultra-magnification as a sequence of in-distribution scale transitions. This design ensures stable inference at arbitrary scales while requiring only a single model. CASR tackles two major bottlenecks: distribution drift across iterations and patch-wise diffusion inconsistencies. The proposed SSAM module aligns structural distributions via superpixel aggregation, preventing error accumulation, while SARM module restores high-frequency textures by enforcing correlation-guided consistency and preserving self-similarity structure through correlation alignment. Despite using only a single model, our approach significantly reduces distribution drift, preserves long-range texture consistency, and achieves superior generalization even at extreme magnification. The source code of our method can be found at
\href{https://github.com/cvwhguo/CASR}{\textcolor{magenta}{https://github.com/cvwhguo/CASR}}.
\end{abstract}    
\section{Introduction}
\label{sec:intro}
Arbitrary-Scale Image Super-Resolution (ASISR) aims to reconstruct high-resolution (HR) images at arbitrary scaling factors from a single low-resolution (LR) input using one unified model. While existing methods~\cite{liif,ciaosr,linf,tsao2024boosting,idm} perform well within their trained scale ranges, they degrade sharply once the inference scale moves beyond this regime. This failure is rooted in cross-scale distribution shift, where the LR-to-HR mapping, texture statistics, and reconstruction priors become inconsistent under large-scale jumps, leading to blur, detail loss, and severe artifacts. Such instability fundamentally limits the practicality and scalability of ASISR in real-world ultra-magnification scenarios.

\begin{figure}[tp]
\centering
\includegraphics[width=0.85\linewidth]{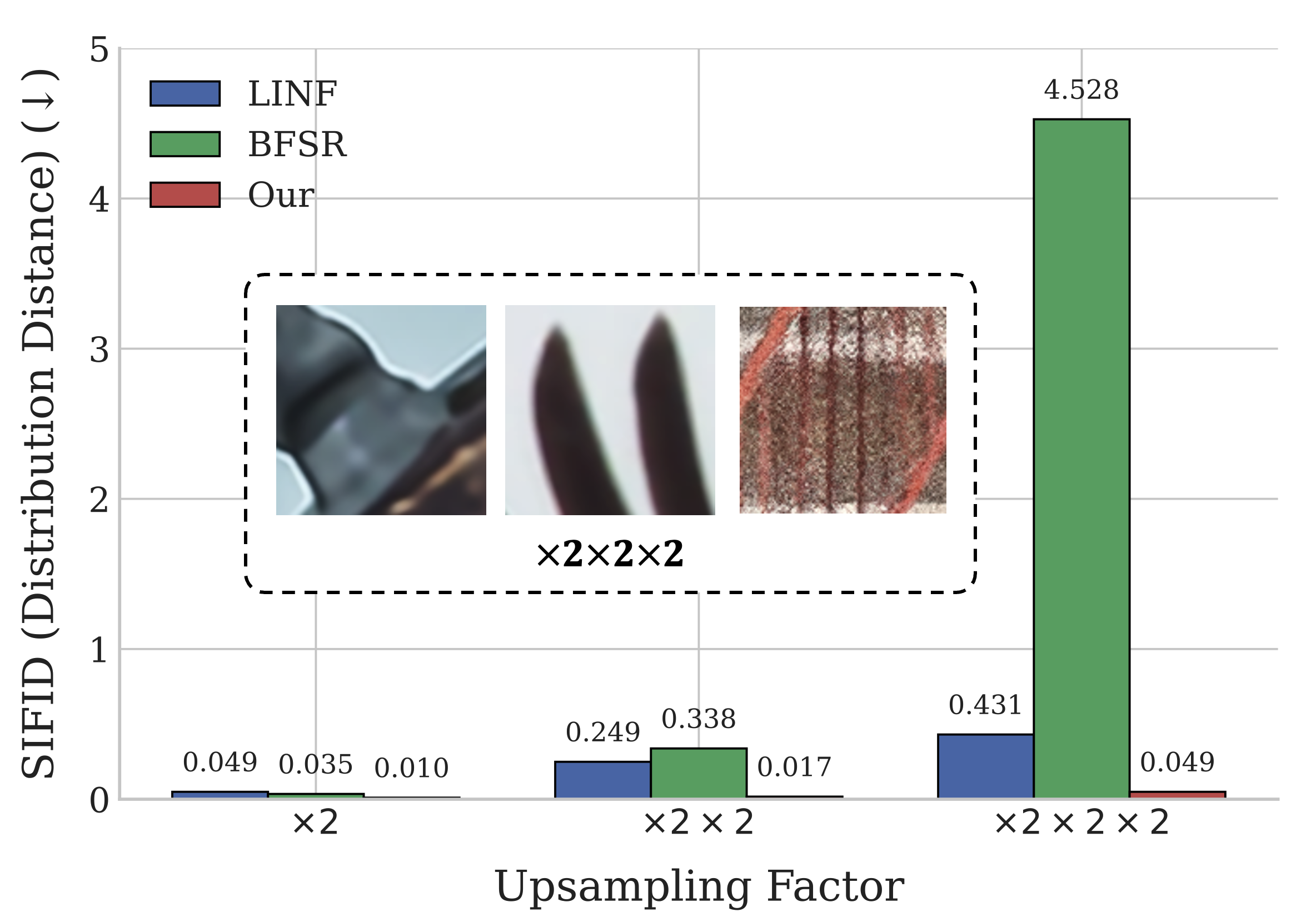} \\
\centering
\caption{Comparison of cyclic cascade stability across different ASISR. The SIFID measures distribution shifts between reconstructed images and the training data during cascading. Our method achieves notably higher distribution stability than others.}\label{fig:iter_data}
\end{figure}

A straightforward strategy to address this extrapolation challenge is to enlarge the training scale range. However, the ill-posed one-to-many mapping in SR becomes increasingly unstable as the scale expands, making optimization intractable and convergence unreliable. Cascading multiple specialized SR networks is another option, but such pipelines incur substantial parameter redundancy, storage overhead, and lack flexibility for dynamically varying scales.

To overcome these limitations, we propose a cyclic reusable single-network architecture that interprets ultra-magnification as a sequence of in-distribution scale transitions. Instead of directly predicting large upscaling factors, which forces the model outside its training distribution, CASR progressively enlarges images by repeatedly applying the same SR model. Each step remains within the learned distribution, ensuring stable inference while significantly reducing complexity and memory usage. This cyclic formulation achieves high-quality reconstruction through gradual refinement, offering an elegant, scalable, and efficient solution for ASISR.

\begin{figure}[tp]
\centering
\includegraphics[scale=1.5]{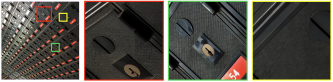} \\
\centering
\caption{This illustrates the texture inconsistency between patches caused by patch-based super-resolution, where identical repeated objects are reconstructed with different texture patterns.}\label{fig:textures}
\vspace{-10pt}
\end{figure}

However, building a robust cyclic SR framework introduces two key challenges. First, recursive application of an SR model inevitably causes distribution drift: intermediate outputs gradually deviate from the training distribution, amplifying residual noise, ringing, and blur through positive feedback. As illustrated in Fig.~\ref{fig:iter_data}, the SIFID \cite{SinGAN2019} metric increases consistently with each iteration, illustrating this drift and the resulting quality degradation. Second, diffusion-based priors have shown remarkable potential in enhancing texture realism, but most diffusion backbones impose strict input resolution constraints. In cyclic ASISR, progressively enlarged images must be processed in smaller patches due to memory limits, followed by reassembly into full-resolution outputs. While overlap blending \cite{bar2023multidiffusion, jimenez2023mixture} partially mitigate boundary artifacts, cross-patch self-similarity, the coherence of textures and repeated structures across adjacent patches, remains difficult to preserve, as shown in Fig.~\ref{fig:textures}.

To address these challenges, we propose a novel super-resolution framework, CASR, consisting of two key components: a Superpixel-based Structural Alignment Module (SSAM) and a Self-similarity Aware Refinement Module (SARM). SSAM stabilizes the cross-scale distribution transitions within the cyclic process by grouping visually similar pixels into homogeneous superpixel regions. This suppresses isolated noise, mitigates edge misalignment, and adaptively controls region granularity to prevent artifacts and over-smoothing. A normalized depth constraint further corrects spatial misalignment during local upsampling, preserving structural consistency across intermediate representations. SARM aims to restore high-frequency textures lost during degradation. Guided by a correlation-guided loss, the network captures repetitive structures, while a cross-attention mechanism leverages global semantics from the low-resolution input to maintain cross-patch consistency.

Our main contributions are summarized as follows:
\begin{itemize}
\item We propose CASR, a simple yet theoretically grounded cyclic SR framework that models ultra-magnification as a sequence of in-distribution scale transitions, fundamentally mitigating cross-scale distribution shift.
\item We design SSAM and SARM to jointly stabilize distribution drift and preserve cross-scale self-similarity, enabling CASR to achieve coherent textures and state-of-the-art performance under extreme magnification.
\end{itemize}

\section{Related Work}
\label{sec:relatedwork}

\subsection{Arbitrary-Scale Super-Resolution}

MetaSR~\cite{metasr} introduced a meta-upscaling module that dynamically generates filter weights based on input coordinates and scaling factors, enabling arbitrary magnification. However, its generalization drops sharply at large scales. LIIF~\cite{liif} adopts implicit neural representation (INR), where an MLP predicts RGB values for queried coordinates from local LR features, allowing extrapolation beyond training scales.Subsequent works integrated generative paradigms such as normalizing flows~\cite{linf,tsao2024boosting} and diffusion models~\cite{idm,kim2024arbitrary} to improve perceptual fidelity. LINF~\cite{linf} first combined normalizing flows with INR for arbitrary-scale SR, while BFSR \cite{tsao2024boosting} introduced conditional learning for further gains. IDM \cite{idm} and Kim \cite{kim2024arbitrary} incorporated diffusion priors, achieving state-of-the-art perceptual results on category-specific datasets. Despite the impressive performance of implicit neural representation methods for ASISR capable of scaling up to $\times 30$—these models still encounter issues with blur or distortion when recovering fine image details in scenarios exceeding $\times 4$.

\begin{figure*}[tp]
\centering
\includegraphics[scale=0.25]{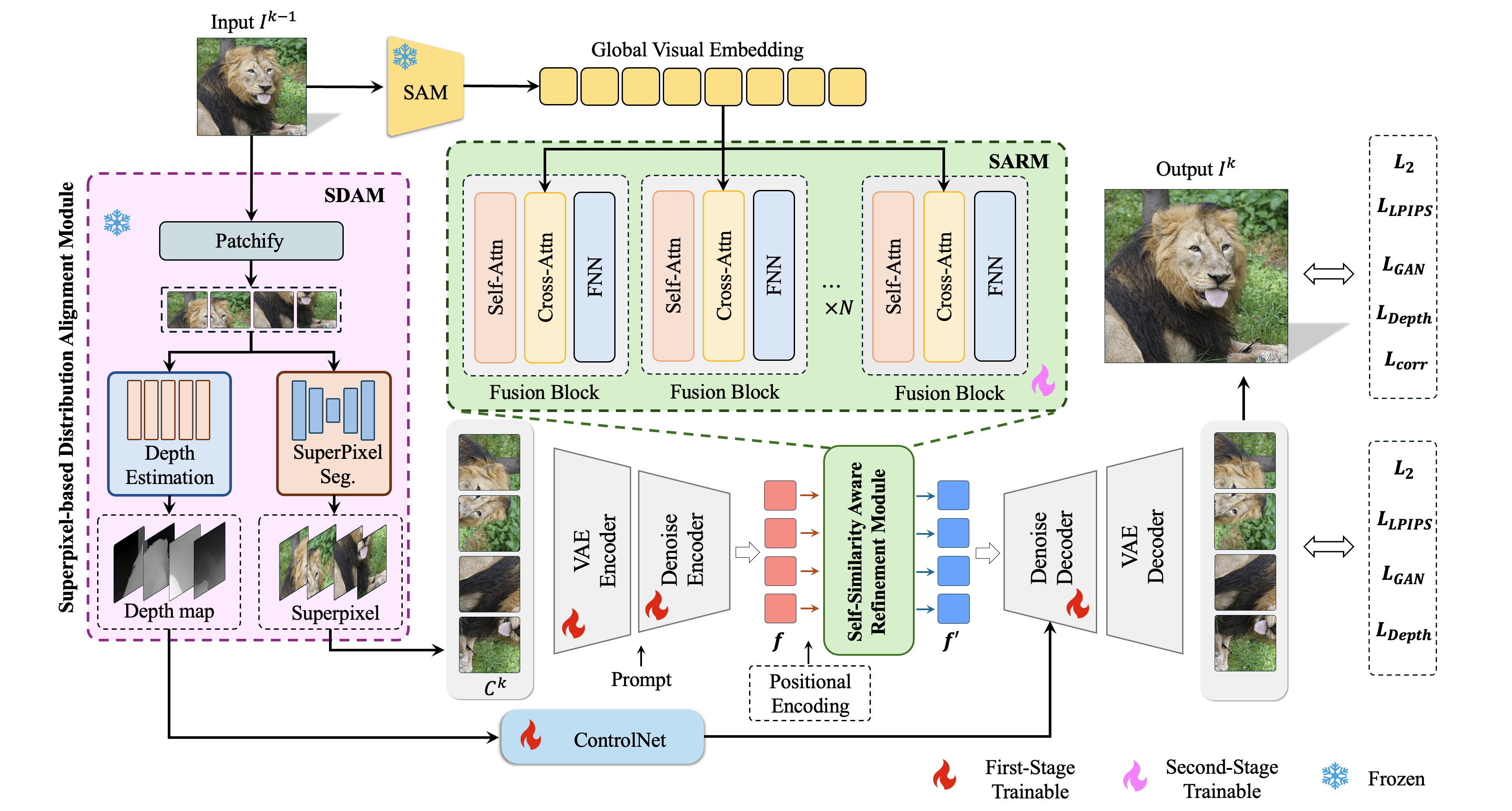} \\
\centering
\caption{Illustration of the proposed \textbf{CASR}. The purple module denotes the SSAM, the green block corresponds to the SARM, and the gray U-Net represents the SR backbone.}\label{fig:method}
\vspace{-1em}
\end{figure*}

\subsection{Large-scale Image Super-Resolution}
Large-scale SR has traditionally been explored under fixed upsampling factors. Many methods attempt to mitigate the ill-posedness of extreme downsampling by training on LR–HR pairs up to $\times16$ or $\times64$, often within restricted semantic domains. For example, PULSE~\cite{pulse} optimizes StyleGAN~\cite{stylegan} latent codes to generate HR faces consistent with LR inputs, while GLEAN~\cite{glean} enhances spatial consistency by conditioning StyleGAN on convolutional features, extending scalability to $\times64$. However, these generative approaches remain domain-specific—typically limited to faces, cats, or other well-structured categories—and struggle to generalize to arbitrary real-world content.

Cascaded SR frameworks~\cite{imagen,dell_e} address scalability by sequentially applying multiple SR models. SR3~\cite{sr3} stacks several $\times4$ diffusion-based models to achieve $\times64$ upsampling. Yet, the mismatch between intermediate outputs and the next model’s training distribution causes performance degradation across stages. CDM~\cite{cascaded} alleviates this issue through noise and blur augmentation but requires training and storing multiple networks, limiting scalability. Such cascaded pipelines are ill-suited for arbitrary-scale SR. 


\section{Method}
Given a low-resolution (LR) image and an arbitrary scaling factor \( s \), our goal is to reconstruct a high-resolution (HR) image at arbitrary magnification. To support ultra-large scaling, we decompose \( s \) into a series of sub-scale factors, each bounded by a predefined maximum scale \( s_{\text{max}} \) used during training: $s = s^1 \times s^2 \times \cdots \times s^k \times \cdots \times s^K$, where \( s^k \leq s_{\text{max}} \) and \( k \in [1, K] \).  
The proposed CASR framework performs \( K \) iterative upsampling steps, where each intermediate result serves as the input for the next. In the \( k \)-th iteration, the upsampling module enlarges \( I^{k-1} \) by a factor of \( s^k \), producing \( I^k \). Starting from \( I^0 \), this process continues until the final HR output \( I^K \) is obtained. As shown in Fig.~\ref{fig:method}, the input \( I^{k-1} \) is first processed by the SSAM and divided into patches. Superpixel and depth maps are extracted and fed into the SR backbone, followed by refinement via the SARM to ensure texture consistency. All enhanced patches are then assembled into the final full-resolution output \( I^k \). 

We next detail the SSAM in Sec. ~\ref{sec:SSAM} and SARM in Sec. ~\ref{sec:SARM}, followed by the overall training strategy in Sec. ~\ref{sec:train}.

\subsection{Superpixel-based Structural Alignment }
\label{sec:SSAM}
During the cyclic super-resolution process, reconstruction artifacts tend to accumulate across iterations, leading to a significant distribution shift. As the SR network repeatedly enhances edges and textures, residual noise, ringing effects, and local blurring are unintentionally introduced, progressively altering the feature statistics of the reconstructed image. Consequently, the input of subsequent iterations deviates from the model’s original training manifold.

To mitigate this problem, we propose the Superpixel-based Structural Alignment Module (SSAM), a novel distribution alignment strategy that suppresses these degradation factors at an early stage while separating the stable structural components from noisy artifacts.

\begin{figure}[t]
\centering
\begin{minipage}[h]{0.95\linewidth}
    \begin{minipage}[h]{1\linewidth}
        \centering
        \begin{minipage}[h]{0.19\linewidth}
            \centering
            \includegraphics[width=1\linewidth]{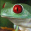}\\  
        \end{minipage}
        \hfill
        \begin{minipage}[h]{0.19\linewidth}
            \centering
            \includegraphics[width=1\linewidth]{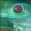}\\    
        \end{minipage}
        \hfill
        \begin{minipage}[h]{0.19\linewidth}
            \centering
            \includegraphics[width=1\linewidth]{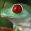}\\
        \end{minipage}
        \hfill
        \begin{minipage}[h]{0.19\linewidth}
            \centering
            \includegraphics[width=1\linewidth]{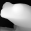}\\
        \end{minipage}
        \hfill
        \begin{minipage}[h]{0.19\linewidth}
            \centering
            \includegraphics[width=1\linewidth]{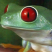}\\
        \end{minipage}
    \end{minipage}
    \vspace{0.005\linewidth}\\
    \begin{minipage}[h]{1\linewidth}
        \centering
        \begin{minipage}[h]{0.19\linewidth}
            \centering
            \includegraphics[width=1\linewidth]{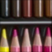}\\  
            \scriptsize{{origin}}
        \end{minipage}
        \hfill
        \begin{minipage}[h]{0.19\linewidth}
            \centering
            \includegraphics[width=1\linewidth]{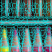}\\    
            \scriptsize{{superpixel-seg.}}
        \end{minipage}
        \hfill
        \begin{minipage}[h]{0.19\linewidth}
            \centering
            \includegraphics[width=1\linewidth]{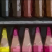}\\
            \scriptsize{{superpixel}}
        \end{minipage}
        \hfill
        \begin{minipage}[h]{0.19\linewidth}
            \centering
            \includegraphics[width=1\linewidth]{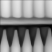}\\
            \scriptsize{{depth}}
        \end{minipage}
        \hfill
        \begin{minipage}[h]{0.19\linewidth}
            \centering
            \includegraphics[width=1\linewidth]{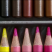}\\
            \scriptsize{{output (ours)}}
        \end{minipage}
    \end{minipage}
\end{minipage}

\centering
\caption{Illustration of the distribution alignment process, where the input image is decomposed into a superpixel representation and a depth map. This decomposition effectively removes artifacts and noise, enabling robust SR.}\label{fig:superpixel}
\vspace{-1.2em}
\end{figure}

\noindent\textbf{Superpixel-based Structural Filtering.}
We first employ a superpixel segmentation strategy to eliminate accumulated artifacts by partitioning the image into coherent and visually homogeneous regions. Superpixels group perceptually similar pixels into compact, uniform segments that approximate sparse coding representations, leading to a smoother and more structured image representation. This process effectively removes cascading artifacts while preserving essential image content. Moreover, the explicit superpixel boundaries facilitate vector-like upsampling through simple nearest-neighbor interpolation.

In our framework, the input is uniformly divided into $n \times n$ superpixel blocks. We design a lightweight fully convolutional Superpixel Segmentation Network (SSN) to predict the soft assignment probabilities of each pixel to its nine neighboring regions.  The SSN is adapted from SuperPixel-FCN \cite{yang2020superpixel}, where we prune redundant channels (reducing convolutional width by 35\%) and apply knowledge distillation from the original model to maintain segmentation accuracy while improving inference efficiency.

The SSN outputs a probability map indicating the likelihood of each pixel belonging to its surrounding superpixels. After bilinear upsampling followed by an \texttt{argmax} operation, we obtain the segmentation mask $P^{k-1}$, where each label corresponds to an individual region. The normalized representation of region $r$ in $P^{k-1}$ is computed as:
\begin{equation}
C^{k-1}_r = \frac{1}{|r|} \sum_{i \in r} I^{k-1}_i,
\end{equation}
where $I^{k-1}_i$ denotes the pixel intensity at position $i$, and $|r|$ is the size of the region.

\vspace{0.3em}
\noindent\textbf{Depth-guided Geometric Constraint.}
Superpixel representations alone may disrupt edge continuity, as segmentation boundaries can misalign with object contours. To preserve geometric integrity, we complement the superpixel image with depth-based structural cues. Unlike edge detection, which is easily corrupted by artifacts, depth estimation provides robust geometric information. We thus incorporate depth maps obtained from the pretrained DepthAnything \cite{yang2024depth} model as auxiliary constraints.

As a result, the original image is decomposed into two complementary and stable representations throughout the iterative cascade:a superpixel image capturing low-frequency content and a structural image preserving high-frequency geometric details. This dual representation effectively suppresses random noise while retaining semantic content and boundary consistency, providing a stable input distribution for the subsequent SR module, as illustrated in Fig. \ref{fig:superpixel}.

\subsection{Self-Similarity Aware Refinement}
\label{sec:SARM}

Due to GPU memory constraints and the fixed input size of diffusion backbones, progressively upscaled images are divided into smaller patches for independent processing and later reassembly. Although overlapping regions can mitigate boundary artifacts,  existing approaches still struggle to fully preserve fine-grained self-similarity—i.e., the ability of local textures or structures to remain consistent across different patches.

This issue involves two challenges:  
(1) How to represent self-similarity and guide the network to focus on repeating patterns, enabling it to recognize and preserve recurring local textures or structures within the image;  
(2) How to provide the network with self-similarity information and embed it into the learning process, so that independently processed patches can share fine-grained texture and semantic information, thereby maintaining global consistency.

\begin{figure}[!ht]
\centering
\centering
\includegraphics[scale=0.15]{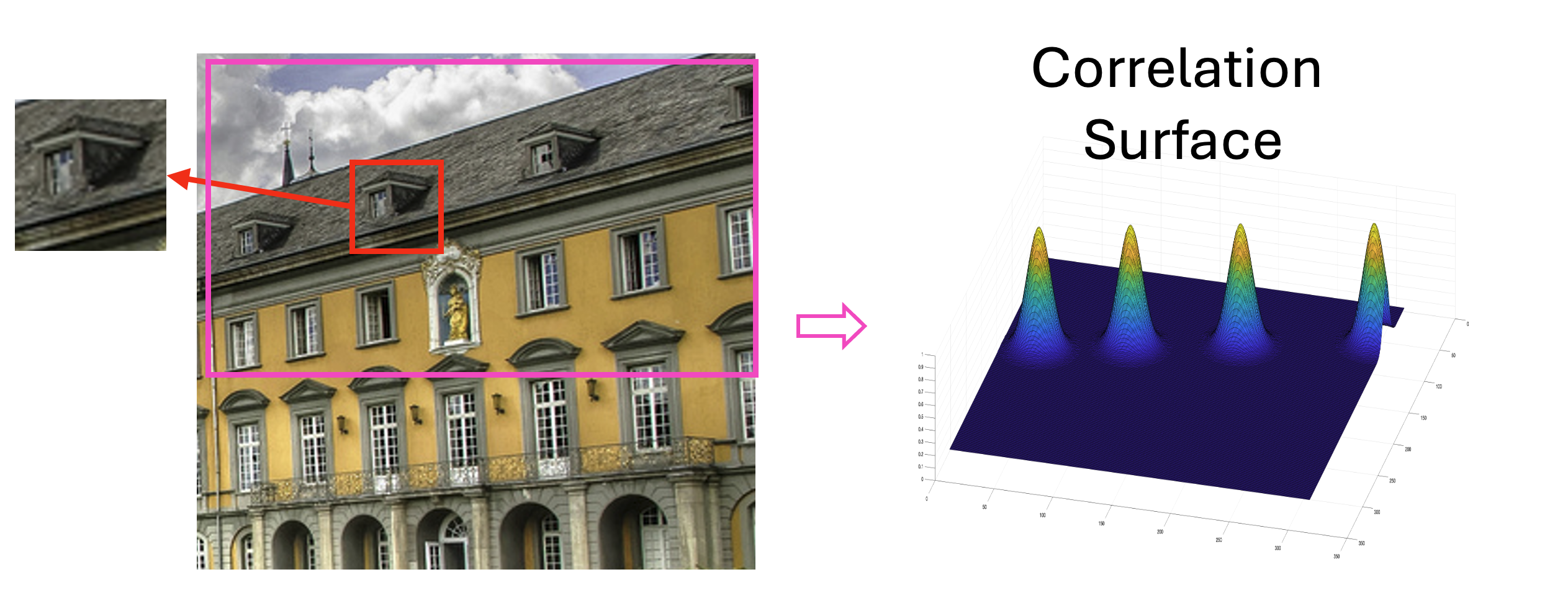} \\
\centering
\caption{Illustration of the local self-similarity computation, where structurally similar regions are assigned higher correlation.}\label{fig:self_corr}
\end{figure}

In general, the self-similarity of an image can be expressed through correlations in deep feature space. Let \(e\) denote the feature map extracted from an image by a pretrained encoder. For any local feature vector \(e_i\), its similarity to the entire image can be written as \(r_i = e_i e^\top\), where \(r_i \in \mathbb{R}^{hw \times 1}\) describes the correlation between the current location and all spatial positions, as show Fig. \ref{fig:self_corr}. Preserving this correlation structure during super-resolution helps maintain the intrinsic self-similarity of the image.
For the reconstructed image \(I^k\) and the ground truth \(I^{\text{gt}}\), we extract their semantic embeddings \(e^k\) and \(e^{\text{gt}}\) using a pretrained SAM ~\cite{kirillov2023segment} encoder, and compute their self-similarity matrices:
\begin{equation}
R^k = e^k (e^k)^\top, \qquad
R^{\text{gt}} = e^{\text{gt}} (e^{\text{gt}})^\top .
\end{equation}
These matrices compactly encode pairwise similarities within each image, providing a robust representation of internal self-similarity.

To incorporate this information into the reconstruction process, we introduce a Self-Similarity Aware Refinement Module (SARM) that enables cross-patch information exchange. Concretely, each patch is independently encoded, and the bottleneck features $f$ at the lowest resolution are cached. Global self-attention is then computed over these cached features—incurring minimal memory overhead due to the small token count—before the globally fused features are independently decoded. Adjacent patches are merged via MultiDiffusion~\cite{bar2023multidiffusion} to produce the final HR output. As shown in Fig.~\ref{fig:method}, the attention mechanism aggregates global information from $f$, allowing each patch to perceive the spatial distribution of patterns across the whole image. To mitigate the loss of global context caused by patch-wise processing, we further extract a global semantic embedding $g$ from the LR image $I^{k-1}$ using SAM, and introduce it via cross-attention. In contrast to \cite{inf_dit}, which emphasizes local neighbor cues, our design explicitly incorporates global semantic context to strengthen cross-patch consistency.
We additionally enforce this self-similarity structure through a correlation-guided objective:
\begin{equation}
L_{\text{corr}} = \left|\left| R^k - R^{\text{gt}} \right|\right|_2 .
\end{equation}
This correlation-guided loss enforces consistent similarity relationships among semantically related regions, ensuring coherent textures and structures in the final output.


\subsection{Training Strategy}
\label{sec:train}
We adopt SD-Turbo \cite{sd_turbo} as the backbone of our framework, which is a single-step diffusion model optimized for fast generation. During fine-tuning, all pretrained parameters are kept frozen, and lightweight adaptation is achieved through LoRA modules applied to both the VAE encoder and the denoising U-Net. To ensure deterministic refinement, the stochastic noise injection process in diffusion sampling is disabled. Given a superpixel-aligned input image, the module performs encoding, denoising, and decoding through a VAE–U-Net–VAE pipeline, while structural control signals from the ControlNet \cite{zhang2023adding} branch are injected into the U-Net decoder to guide structure-aware reconstruction. The final output is a high-quality, detail-enhanced high-resolution image.

\vspace{0.3em}
\noindent\textbf{Two-stage Training.}
CASR is trained in two stages: a \emph{super-resolution stage} and a \emph{self-similarity stage}.  
In the first stage, the SD-Turbo backbone is fine-tuned while omitting the SARM, focusing on high-quality reconstruction with both perceptual and structural fidelity. The reconstruction loss is:
\begin{equation}
L_{\text{rec}} = \lambda_1 L_2 + \lambda_2 L_{\text{LPIPS}} + \lambda_3 L_{\text{GAN}}.
\end{equation}
To better exploit geometric cues, a depth consistency loss is introduced. Given depth maps $d^k$ and $d^{\text{gt}}$ from DepthAnything \cite{yang2024depth}, both normalized to $[0, 1]$, the alignment loss is:
\begin{equation}
L_{\text{depth}} = \left\| \text{Norm}(d^k) - \text{Norm}(d^{\text{gt}}) \right\|_2.
\end{equation}
The total loss for the first stage is:
\begin{equation}
L_{\text{total}_1} = L_{\text{rec}} + \lambda_4 L_{\text{depth}}.
\end{equation}

In the second stage, the backbone and ControlNet are frozen, and the global fusion module is trained with an additional correlation term:
\begin{equation}
L_{\text{total}_2} = L_{\text{rec}} + \lambda_4 L_{\text{depth}} + \lambda_5 L_{\text{corr}},
\end{equation}
which enhances self-similarity of the reconstructed image.

\begin{table*}[t]
    \caption{Comparison with ASISR methods on the DIV8K synthetic dataset, with the best results in \textbf{bold}. The $\times 4 \times 3 \times 1.5$ column evaluates all methods under progressive upsampling. \label{table:div8K-large}}
    \scriptsize
    \centering
    \resizebox{0.72\textwidth}{!}{
    \begin{tabular}{  l | c c c c | c c c c | c c c c}
        \toprule
        \multirow{3}{*}{Method}  &  \multicolumn{12}{c}{\textbf{DIV8K}}\\
        \multirow{2}{*}{} & \multicolumn{4}{c|}{$\times$8} & \multicolumn{4}{c|}{$\times$12} & \multicolumn{4}{c}{$\times$18}  \\

        \multirow{2}{*}{} 
        & \textbf{LPIPS}$\downarrow$ 
        & \textbf{MUSIQ}$\uparrow$ 
        & \textbf{NIQE}$\downarrow$ 
        & \textbf{PI}$\downarrow$ 
        & \textbf{LPIPS}$\downarrow$ 
        & \textbf{MUSIQ}$\uparrow$ 
        & \textbf{NIQE}$\downarrow$ 
        & \textbf{PI}$\downarrow$ 
        & \textbf{LPIPS}$\downarrow$ 
        & \textbf{MUSIQ}$\uparrow$ 
        & \textbf{NIQE}$\downarrow$ 
        & \textbf{PI}$\downarrow$ \\

        \midrule
        LINF \cite{linf}  & 0.442&26.01&10.11&8.93&0.528&19.42&11.43&9.87&0.578&17.24&13.37&10.89\\
        BFSR \cite{tsao2024boosting}  & 0.399&24.30&8.29&7.80&0.500&18.43&10.70&9.18&0.561&17.06&14.24&11.06\\
        IDM \cite{idm} & 0.486 & 24.11 & 7.23 & 6.46 & 0.604 & 23.42 & 7.98 & 6.87 & 0.656 & 23.75 & 7.82 & 6.87 \\
        Kim  \cite{kim2024arbitrary} & 0.491 & 23.54 & 8.24 & 8.84 & 0.621 & 21.69 & 8.80 & 7.38 & 0.685 & 20.06 & 8.34 & 7.72 \\
        LIIF \cite{liif}+Diff &0.411&28.99&9.32&8.42&0.496&20.25&10.86&9.44&0.550&17.58&12.17&10.21\\
        CiaoSR \cite{ciaosr}+Diff& 0.408&30.94&9.15&8.34&0.493&20.89&10.64&9.33&0.545&17.61&11.93&10.10\\
        \midrule
        CASR & \textbf{0.363} & \textbf{53.63} & \textbf{5.66} & \textbf{5.07} & \textbf{0.403} & \textbf{53.82} & \textbf{5.47} & \textbf{4.89} & \textbf{0.450} & \textbf{51.44} & \textbf{6.01} & \textbf{5.24} \\

        \midrule
        \multirow{2}{*}{Method} & \multicolumn{4}{c|}{$\times$24} & \multicolumn{4}{c|}{$\times$30} & \multicolumn{4}{c}{$\times 4 \times 3 \times 1.5$}  \\
        \multirow{2}{*}{} 
        & \textbf{LPIPS}$\downarrow$ 
        & \textbf{MUSIQ}$\uparrow$ 
        & \textbf{NIQE}$\downarrow$ 
        & \textbf{PI}$\downarrow$ 
        & \textbf{LPIPS}$\downarrow$ 
        & \textbf{MUSIQ}$\uparrow$ 
        & \textbf{NIQE}$\downarrow$ 
        & \textbf{PI}$\downarrow$ 
        & \textbf{LPIPS}$\downarrow$ 
        & \textbf{MUSIQ}$\uparrow$ 
        & \textbf{NIQE}$\downarrow$ 
        & \textbf{PI}$\downarrow$ \\

        \midrule
        LINF \cite{linf}  & 0.608&16.42&15.34&11.86&0.625&16.36&16.32&12.28&0.640&19.04&9.39&6.45\\
        BFSR \cite{tsao2024boosting}  & 0.594&16.42&16.19&12.05&0.611&16.49&16.71&12.21&0.772&17.50&11.07&7.26\\
        IDM \cite{idm} &0.710 & 23.76 & 8.03 & 7.21 & 0.705 & 23.84 & 7.96 & 7.33 & 0.608 & 25.06 & 8.62 & 7.80 \\
        Kim \cite{kim2024arbitrary} & 0.796 & 19.45 & 8.57 & 8.14 & 0.709 & 20.06 & 8.48 & 8.29 & 0.604 & 23.19 & 8.72 & 8.09 \\
        LIIF \cite{liif}+Diff &0.582&16.55&13.49&10.86&0.603&16.16&15.69&11.85&0.535 & 20.27 & 11.93 & 9.19 \\
        CiaoSR \cite{ciaosr}+Diff& 0.579&16.35&13.08&10.66&0.603&15.95&15.70&11.87&0.503 & 21.06 & 10.98 & 9.57 \\
        \midrule
        CASR & \textbf{0.495} & \textbf{41.42} & \textbf{6.93} & \textbf{6.18} & \textbf{0.501} & \textbf{41.76} & \textbf{6.98} & \textbf{6.09} & \textbf{0.450} & \textbf{51.44} & \textbf{6.01} & \textbf{5.24} \\

        \bottomrule
    \end{tabular}

    }
\end{table*}

\begin{figure*}[th]
\centering
\begin{minipage}[h]{0.7\linewidth}
    \begin{minipage}[h]{1\linewidth}
        \begin{minipage}[h]{0.233\linewidth}
            \begin{minipage}[h]{1\linewidth}
                \includegraphics[width=1\linewidth, height=1.05\linewidth]{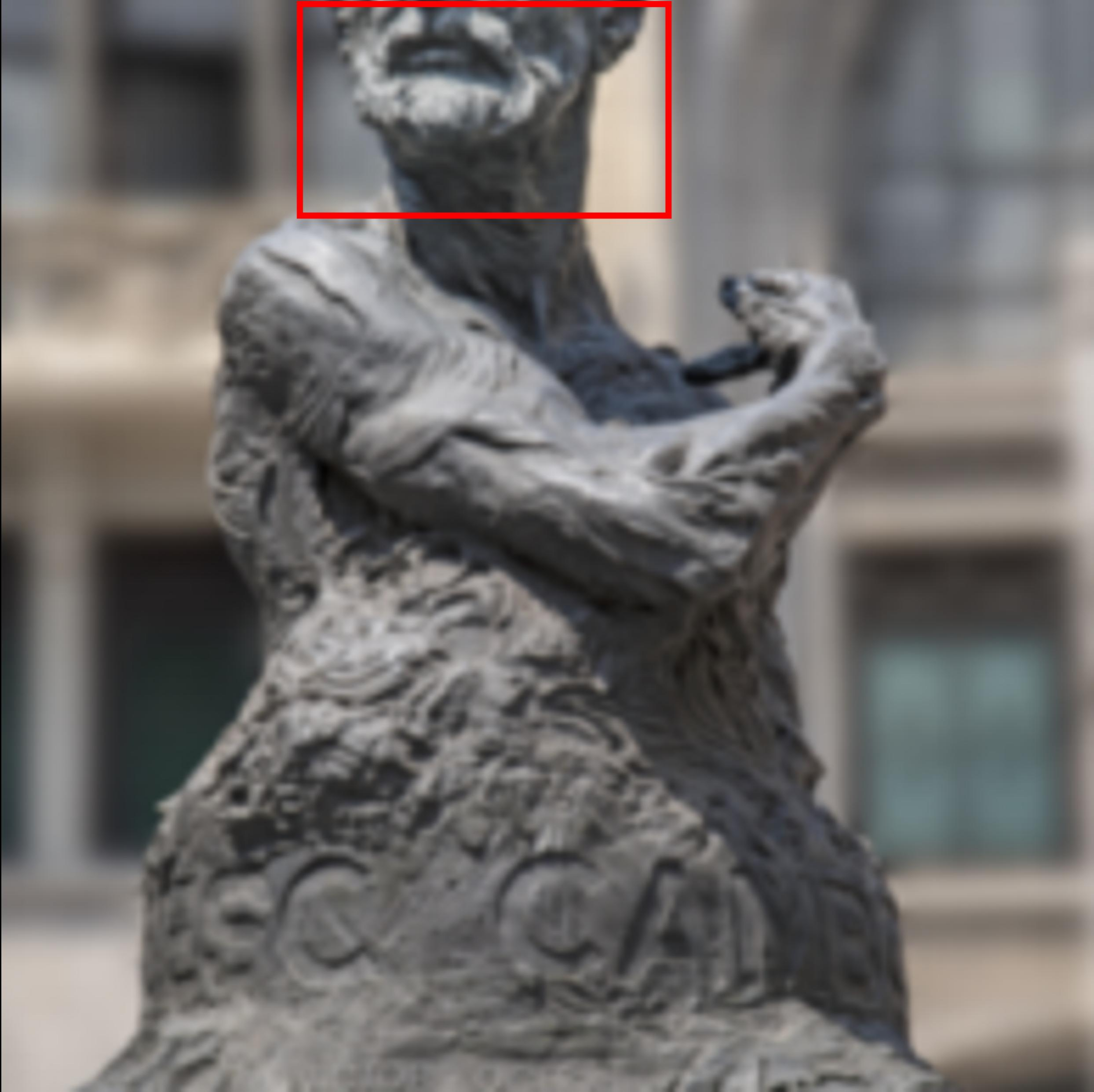} \\
                \centering
                \scriptsize{``DIV8K-1493" $\times 24$}\\
            \end{minipage}
        \end{minipage}
        \hfill
        \begin{minipage}[h]{0.18\linewidth}
            \begin{minipage}[h]{1\linewidth}
                \includegraphics[width=1\linewidth, height=0.6\linewidth]{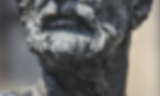} \\
                \centering
                \scriptsize{Bicubic}\\
            \end{minipage}\\     
            \vspace{0.01\linewidth}
            \begin{minipage}[h]{1\linewidth}
                \includegraphics[width=1\linewidth, height=0.6\linewidth]{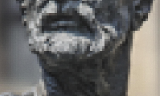} \\
                \centering
                \scriptsize{Kim}
            \end{minipage}\\
        \end{minipage}
        \hfill
        \begin{minipage}[h]{0.18\linewidth}
            \begin{minipage}[h]{1\linewidth}
                \includegraphics[width=1\linewidth, height=0.6\linewidth]{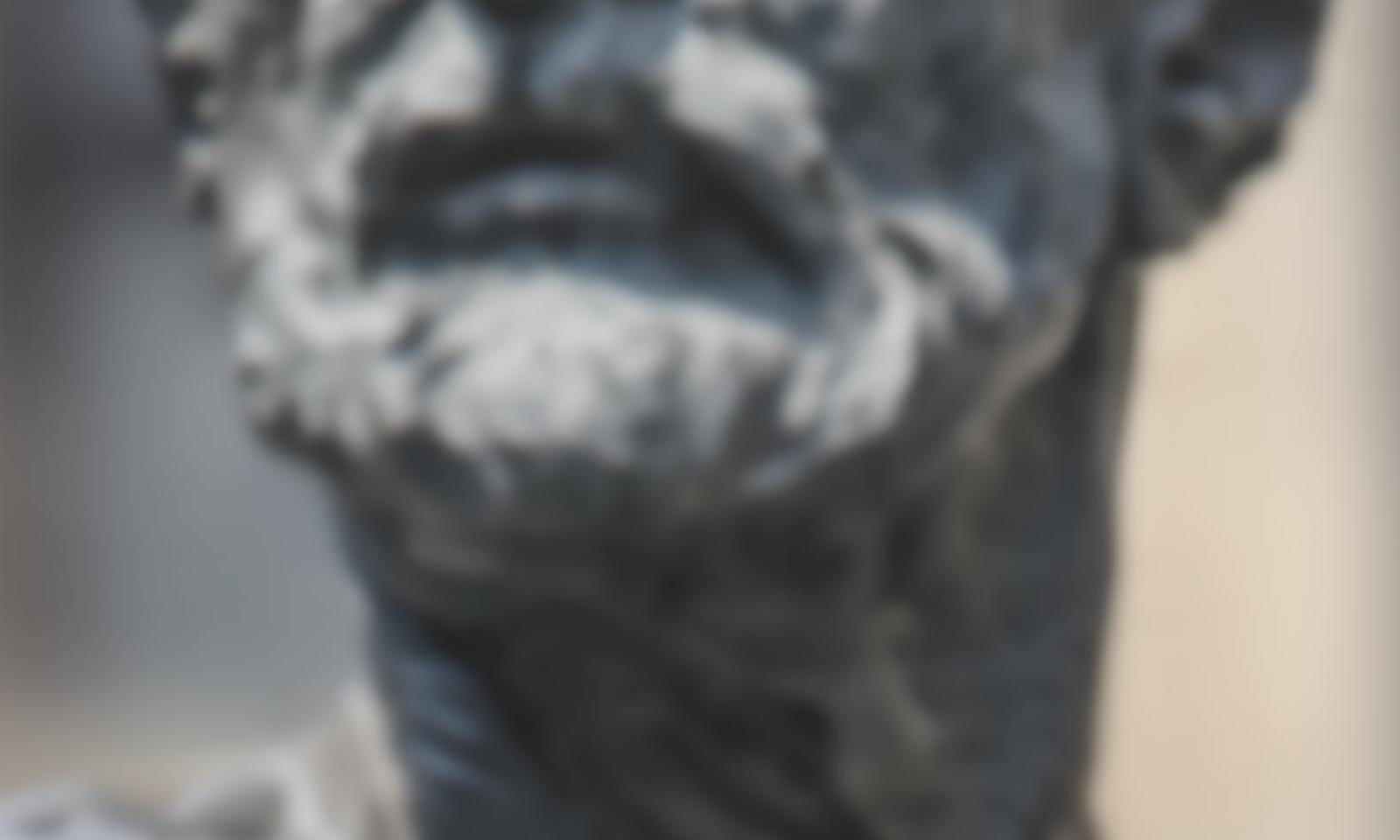} \\
                \centering
                \scriptsize{LINF}\\
            \end{minipage}\\     
            \vspace{0.01\linewidth}
            \begin{minipage}[h]{1\linewidth}
                \includegraphics[width=1\linewidth, height=0.6\linewidth]{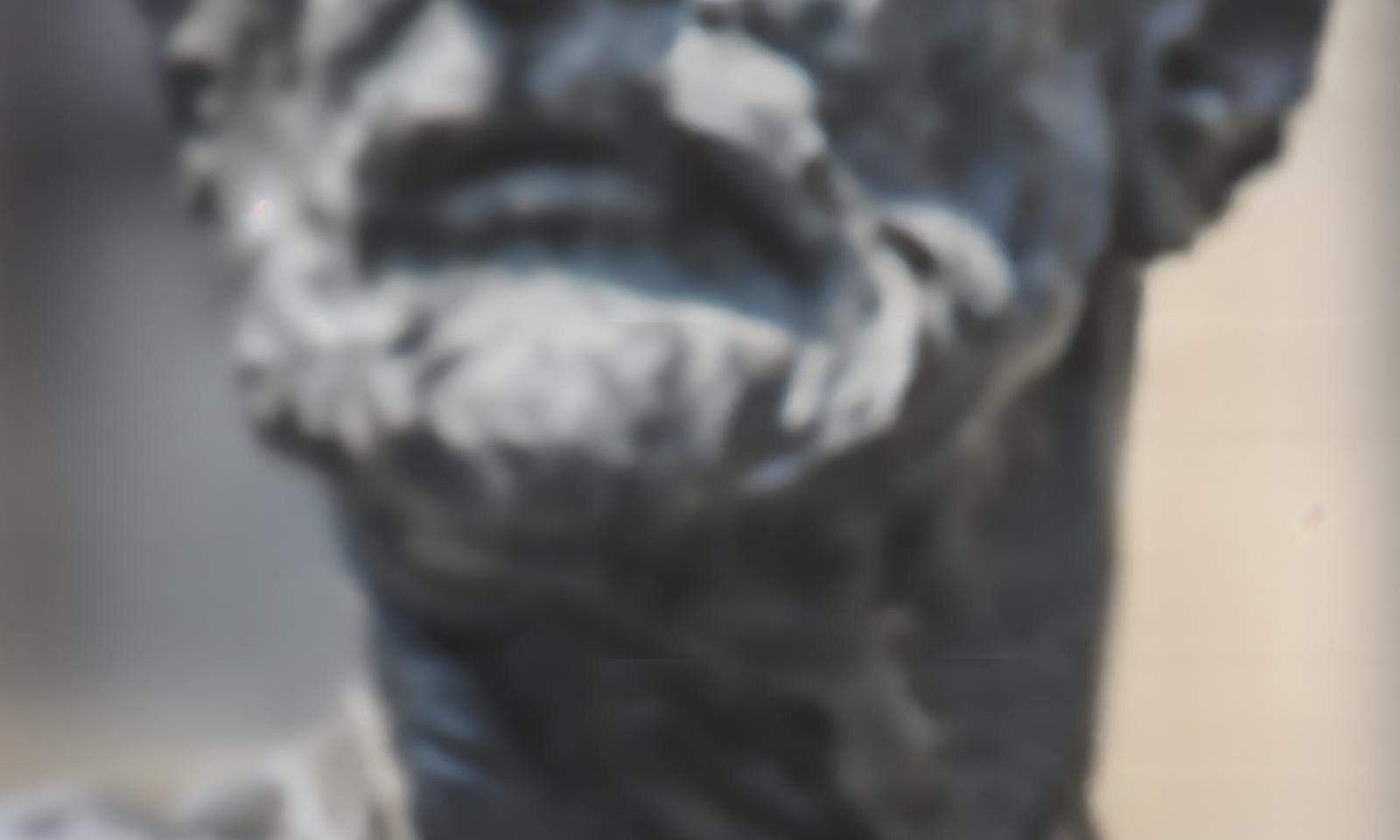} \\
                \centering
                \scriptsize{LIIF+Diff}\\
            \end{minipage}\\
        \end{minipage}
        \hfill
        \begin{minipage}[h]{0.18\linewidth}
            \begin{minipage}[h]{1\linewidth}
                \includegraphics[width=1\linewidth, height=0.6\linewidth]{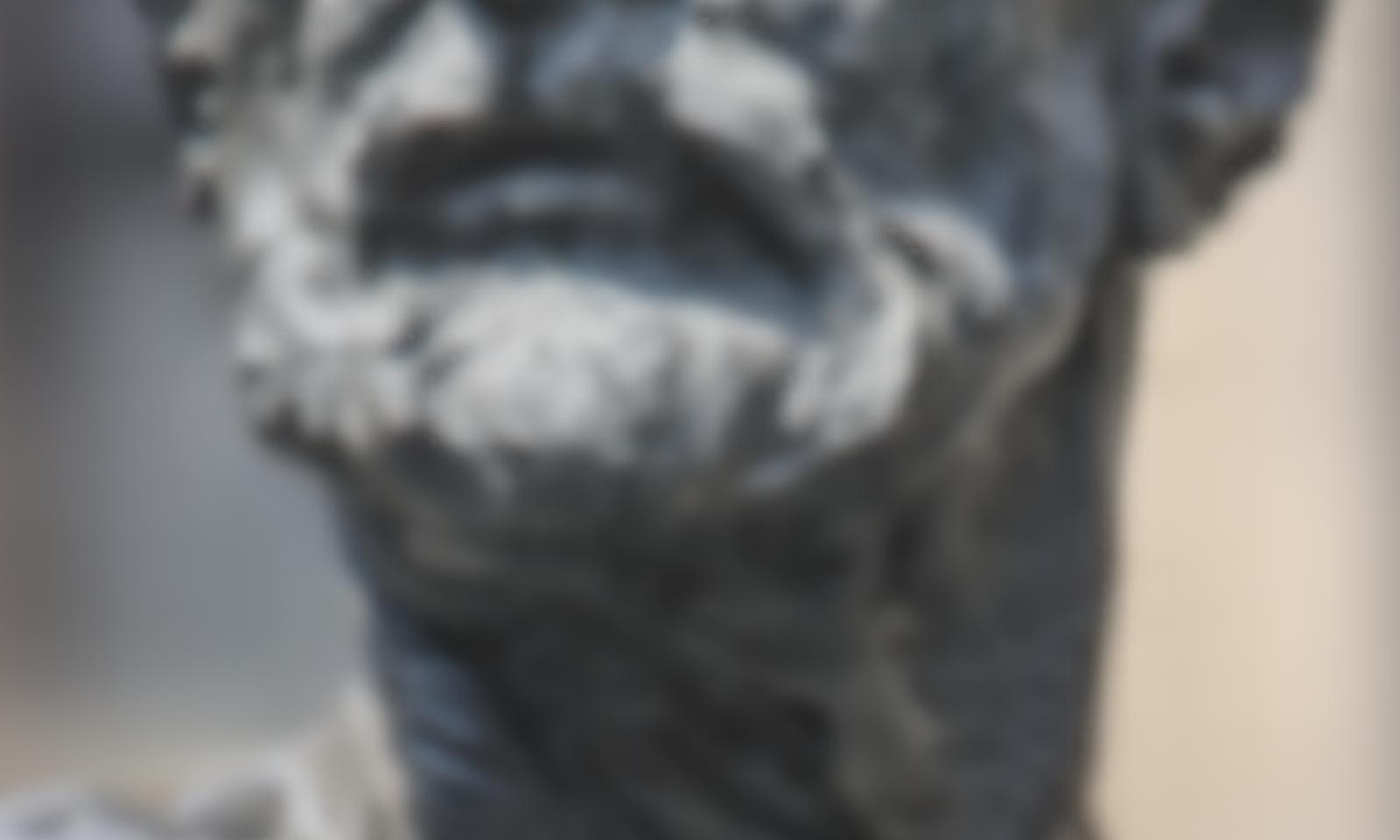} \\
                \centering
                \scriptsize{BFSR}\\
            \end{minipage}\\     
            \vspace{0.01\linewidth}
            \begin{minipage}[h]{1\linewidth}
                \includegraphics[width=1\linewidth, height=0.6\linewidth]{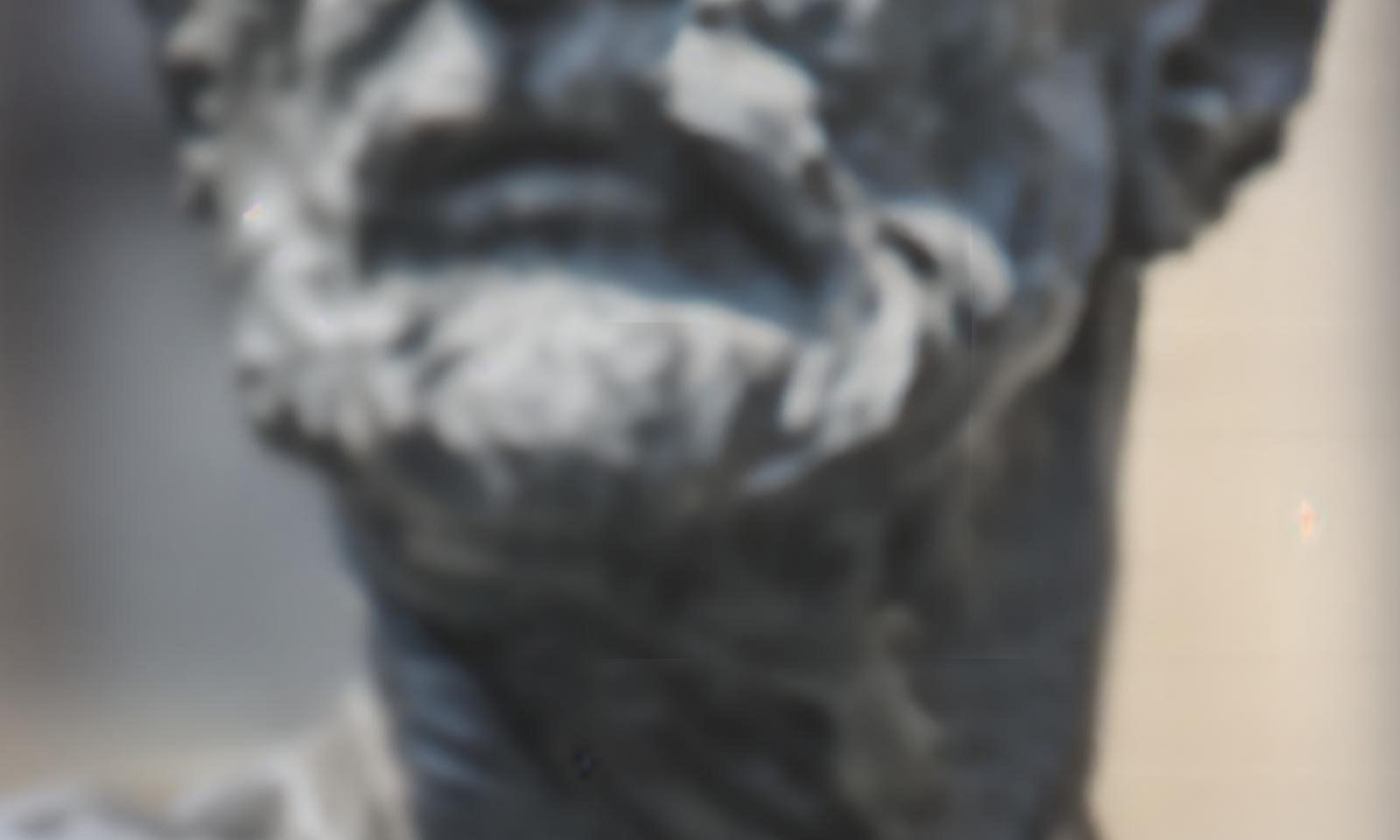} \\
                \centering
                \scriptsize{CiaoSR+Diff}
            \end{minipage}\\
        \end{minipage}
        \hfill
        \begin{minipage}[h]{0.18\linewidth}
            \begin{minipage}[h]{1\linewidth}
                \includegraphics[width=1\linewidth, height=0.6\linewidth]{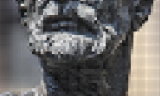} \\
                \centering
                \scriptsize{IDM}\\
            \end{minipage}\\     
            \vspace{0.01\linewidth}
            \begin{minipage}[h]{1\linewidth}
                \includegraphics[width=1\linewidth, height=0.6\linewidth]{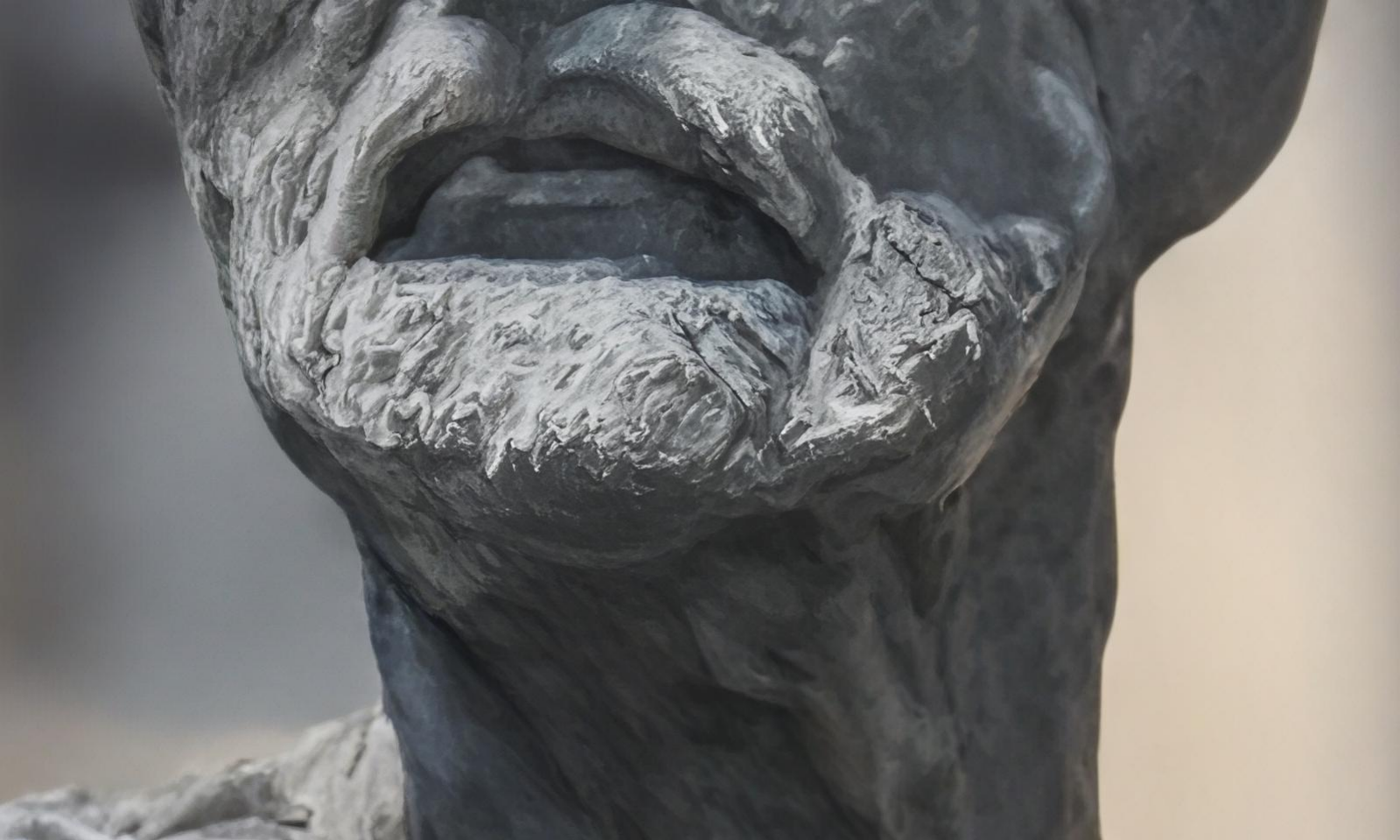} \\
                \centering
                \scriptsize{CASR}
            \end{minipage}\\
        \end{minipage}
    \end{minipage}
    \vspace{0.005\linewidth}

    \begin{minipage}[h]{1\linewidth}
        \begin{minipage}[h]{0.233\linewidth}
            \begin{minipage}[h]{1\linewidth}
                \includegraphics[width=1\linewidth, height=1.05\linewidth]{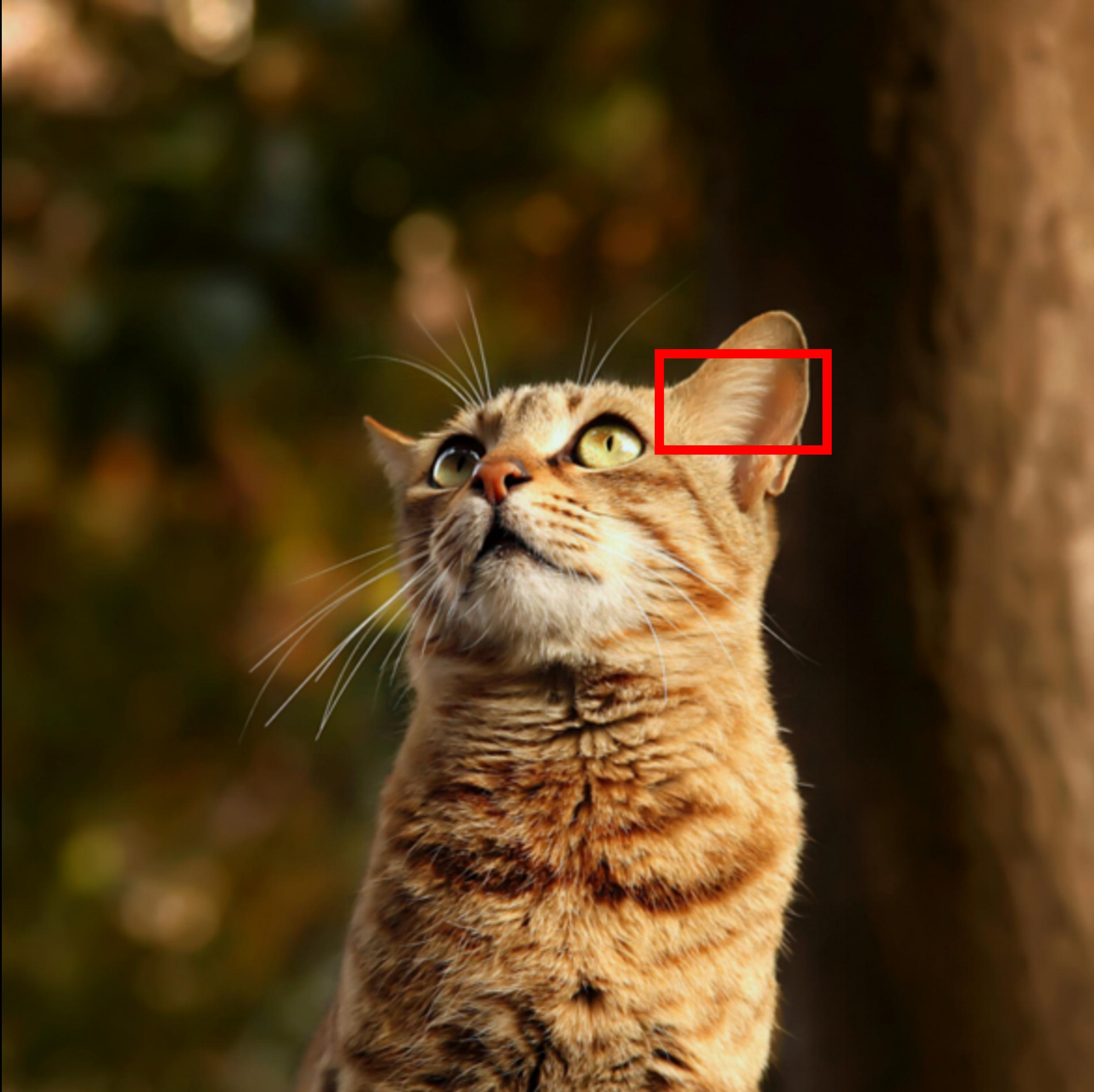} \\
                \centering
                \scriptsize{``DIV8K-1459" $\times 8$}\\
            \end{minipage}
        \end{minipage}
        \hfill
        \begin{minipage}[h]{0.18\linewidth}
            \begin{minipage}[h]{1\linewidth}
                \includegraphics[width=1\linewidth, height=0.6\linewidth]{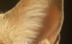} \\
                \centering
                \scriptsize{Bicubic}\\
            \end{minipage}\\     
            \vspace{0.01\linewidth}
            \begin{minipage}[h]{1\linewidth}
                \includegraphics[width=1\linewidth, height=0.6\linewidth]{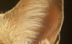} \\
                \centering
                \scriptsize{Kim}
            \end{minipage}\\
        \end{minipage}
        \hfill
        \begin{minipage}[h]{0.18\linewidth}
            \begin{minipage}[h]{1\linewidth}
                \includegraphics[width=1\linewidth, height=0.6\linewidth]{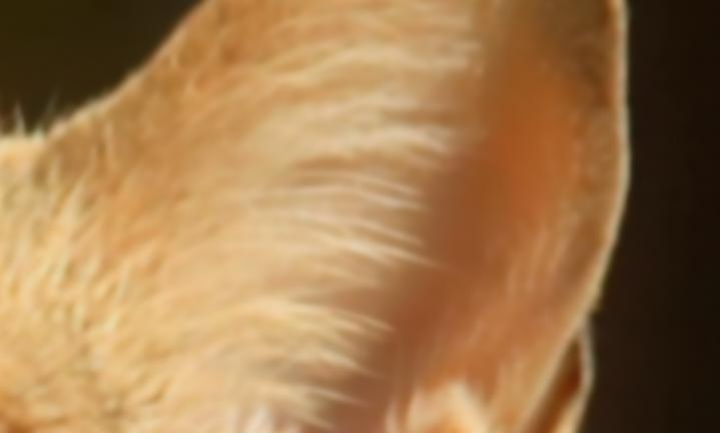} \\
                \centering
                \scriptsize{LINF}\\
            \end{minipage}\\     
            \vspace{0.01\linewidth}
            \begin{minipage}[h]{1\linewidth}
                \includegraphics[width=1\linewidth, height=0.6\linewidth]{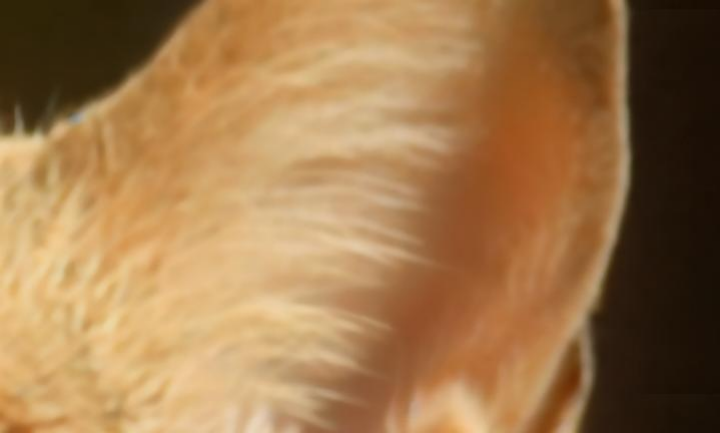} \\
                \centering
                \scriptsize{LIIF+Diff}\\
            \end{minipage}\\
        \end{minipage}
        \hfill
        \begin{minipage}[h]{0.18\linewidth}
            \begin{minipage}[h]{1\linewidth}
                \includegraphics[width=1\linewidth, height=0.6\linewidth]{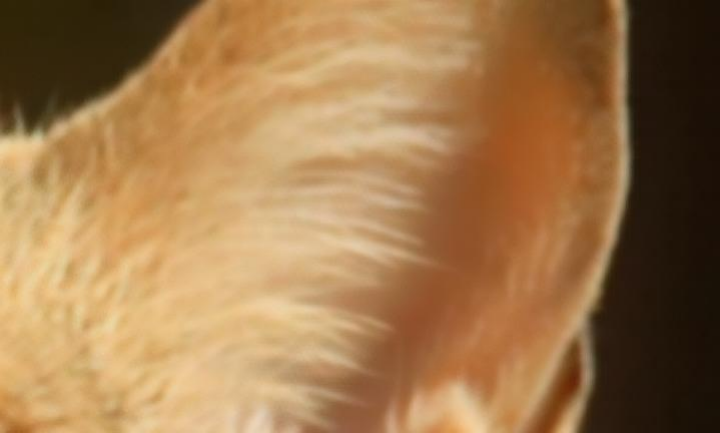} \\
                \centering
                \scriptsize{BFSR}\\
            \end{minipage}\\     
            \vspace{0.01\linewidth}
            \begin{minipage}[h]{1\linewidth}
                \includegraphics[width=1\linewidth, height=0.6\linewidth]{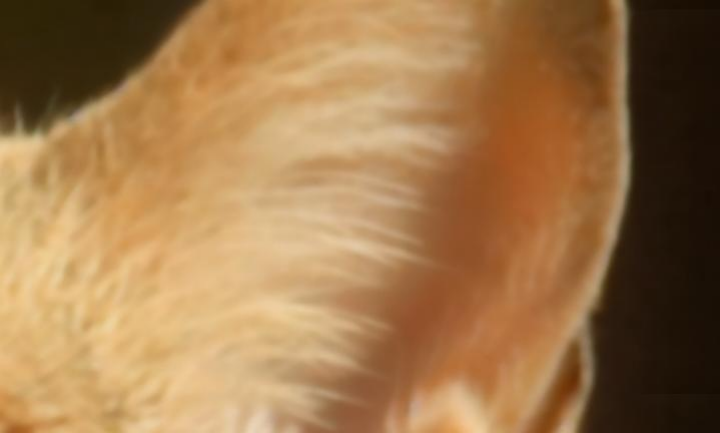} \\
                \centering
                \scriptsize{CiaoSR+Diff}
            \end{minipage}\\
        \end{minipage}
        \hfill
        \begin{minipage}[h]{0.18\linewidth}
            \begin{minipage}[h]{1\linewidth}
                \includegraphics[width=1\linewidth, height=0.6\linewidth]{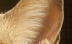} \\
                \centering
                \scriptsize{IDM}\\
            \end{minipage}\\     
            \vspace{0.01\linewidth}
            \begin{minipage}[h]{1\linewidth}
                \includegraphics[width=1\linewidth, height=0.6\linewidth]{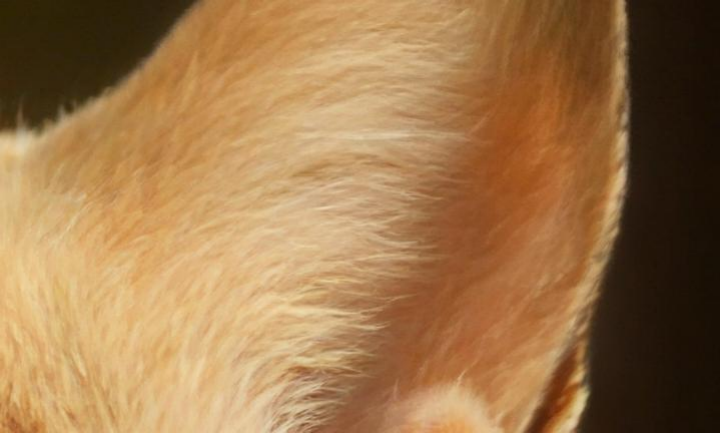} \\
                \centering
                \scriptsize{CASR}
            \end{minipage}\\
        \end{minipage}
    \end{minipage}

\end{minipage}

\centering
\caption{Qualitative comparison with different methods on the DIV8K dataset. For large-scale super-resolution, our method reconstructs more realistic statue textures and finer fur details on the cat's ears.
\label{fig:div8k}} 
\vspace{-1em}
\end{figure*}


\section{Experiments}

\subsection{Training Datasets and Metrics}

Following prior works \cite{hat2023, rcan2018, 2017ntire, eqsr}, we use the DF2K dataset \cite{2017ntire_datasets} for training and synthetically generate low-resolution (LR) images via bicubic downsampling.

For perceptual evaluation, we adopt LPIPS \cite{lpips} along with no-reference image quality assessment metrics, including MUSIQ \cite{musiq}, NIQE \cite{NIQE}, and PI \cite{pi}. For real-world datasets, since ground-truth references are unavailable, LPIPS is excluded from the evaluation.

\subsection{Testing Datasets}
We evaluate our method on three types of datasets: synthetic, real-world, and face datasets. For synthetic evaluation, we adopt the last 100 HR images from the DIV8K dataset \cite{div8k}. The LR inputs are generated by bicubic downsampling of the HR images. For real-world evaluation, we use the RealSR dataset \cite{RealSR}, which contains images captured by two different cameras with complex and authentic degradations. We further evaluate our method on the CelebA-HQ dataset \cite{celeba_hq} following the settings of diffusion-based ASISR methods, including IDM \cite{idm} and Kim \cite{kim2024arbitrary}. Specifically, 100 face images with a resolution of $128 \times 128$ are randomly selected for perceptual evaluation.

\subsection{Implementation Details} \label{sec:Implementation Details}

In all experiments, the maximum upsampling factor \( s_{\text{max}} \) is set to 4. The overall training consists of two stages. In the first stage, we freeze the Structural Alignment Module and train the SR backbone for 10K iterations with a batch size of 32 and a learning rate of \(2 \times 10^{-5}\). Each low-resolution input is first processed by the Structural Alignment Module to produce \(512 \times 512\) superpixel and depth maps, which are then fed into the backbone. During fine-tuning, the LoRA rank parameters are set to 16 for the VAE encoder and 32 for the diffusion U-Net.  In the second stage, we freeze both the Structural Alignment Module and backbone networks, training only the SARM on \(1024 \times 1024\) images divided into four patches, with a batch size of 8, and the same learning rate. Text prompts for the diffusion backbone are dynamically extracted using RAM~\cite{ram} at each upsampling scale. All models are trained on four NVIDIA A6000 GPUs.

\subsection{Comparisons with State-of-the-Art}
We conduct comprehensive comparisons with several sota ASISR methods. These include perceptual quality-driven methods such as LINF \cite{linf}, BFSR \cite{tsao2024boosting}, IDM \cite{idm}, and Kim \cite{kim2024arbitrary}. Since the official checkpoint of Kim \cite{kim2024arbitrary} has not been publicly released, we re-implemented the method according to the descriptions provided in the original paper. In addition, we enhanced LIIF\cite{liif} and CiaoSR\cite{ciaosr} by integrating a \emph{diffusion-based post-processing module}, resulting in the improved variants \emph{LIIF+Diff} and \emph{CiaoSR+Diff}. For high-resolution inference, all methods operate on \(512 \times 512\) patches, using a 64-pixel overlap to ensure seamless boundary blending during stitching.

\begin{table*}[tbh]
    \scriptsize
    \setlength\tabcolsep{3.5pt}
    \centering
    \caption{Comparison with ASISR methods on real-world datasets, with the best results in \textbf{bold}. Our approach archives consistently superior performance over others, showcasing strong generalization in real-world image. } \label{table:real-world-large}
    
    \resizebox{0.85\textwidth}{!}{
    \begin{tabular}{  l | c c c | c c c | c c c | c c c | c c c}      
        \toprule
        \multirow{3}{*}{Method}  &  \multicolumn{15}{c}{\textbf{RealSR}}\\
        \multirow{2}{*}{} & \multicolumn{3}{c|}{$\times$8} & \multicolumn{3}{c|}{$\times$12} & \multicolumn{3}{c|}{$\times$18} & \multicolumn{3}{c|}{$\times$24} & \multicolumn{3}{c}{$\times$30}  \\

        \multirow{2}{*}{} 
        & \textbf{MUSIQ}$\uparrow$ 
        & \textbf{NIQE}$\downarrow$  
        & \textbf{PI}$\downarrow$  
        & \textbf{MUSIQ}$\uparrow$  
        & \textbf{NIQE}$\downarrow$  
        & \textbf{PI}$\downarrow$  
        & \textbf{MUSIQ}$\uparrow$   
        & \textbf{NIQE}$\downarrow$  
        & \textbf{PI}$\downarrow$ 
        & \textbf{MUSIQ}$\uparrow$ 
        & \textbf{NIQE}$\downarrow$  
        & \textbf{PI}$\downarrow$  
        & \textbf{MUSIQ}$\uparrow$ 
        & \textbf{NIQE}$\downarrow$  
        & \textbf{PI}$\downarrow$ \\

        \midrule
        LINF \cite{linf}&19.58&11.50&9.96&17.22&12.90&10.83&16.74&14.59&11.71&17.26&15.70&12.29&18.29&16.26&12.56\\
        BFSR\cite{tsao2024boosting} &18.27 &	9.91 &	8.96	 &16.56 &	12.21 &	10.19	 &16.28	 &15.51 &	11.79 &	16.88 &	16.71 &	12.31 &	18.04 &	16.85 &	12.27\\
        IDM \cite{idm} & 28.69 & 7.50 & 7.15 & 26.56 & 8.14 & 7.67 & 33.79 & 8.00 & 7.52 & 31.68 & 8.17 & 7.47 & 28.22 & 8.35 & 7.44 \\
        Kim \cite{kim2024arbitrary} & 26.39 & 8.09 & 6.94 & 25.16 & 7.83 & 7.85 & 25.32 & 8.51 & 8.42 & 30.59 & 8.58 & 8.46 & 27.43 & 8.56 & 8.43 \\
        LIIF \cite{liif} + Diff & 19.83&10.82&9.36&16.91&12.32&10.34&16.14&13.50&10.90&16.47&13.90&11.12&17.39&14.00&11.12\\
        CiaoSR \cite{ciaosr} +Diff & 20.53&10.62&9.25&17.14&12.04&10.19&16.22&12.99&10.64&16.24&13.23&10.74&17.35&13.26&10.75\\
        \midrule
        CASR & \textbf{53.50} & \textbf{6.81} & \textbf{5.71} & \textbf{49.42} & \textbf{6.73} & \textbf{5.73} & \textbf{44.03} & \textbf{7.56} & \textbf{6.34} & \textbf{40.35} & \textbf{7.80} & \textbf{6.65} & \textbf{37.84} & \textbf{7.81} & \textbf{6.73} \\

        \bottomrule
    \end{tabular}
    }
\end{table*}

\begin{figure*}[t]
\centering
\begin{minipage}[h]{0.85\linewidth}
    \begin{minipage}[h]{1\linewidth}
        \centering
        \begin{minipage}[h]{0.115\linewidth}
            \centering
            \includegraphics[width=1\linewidth]{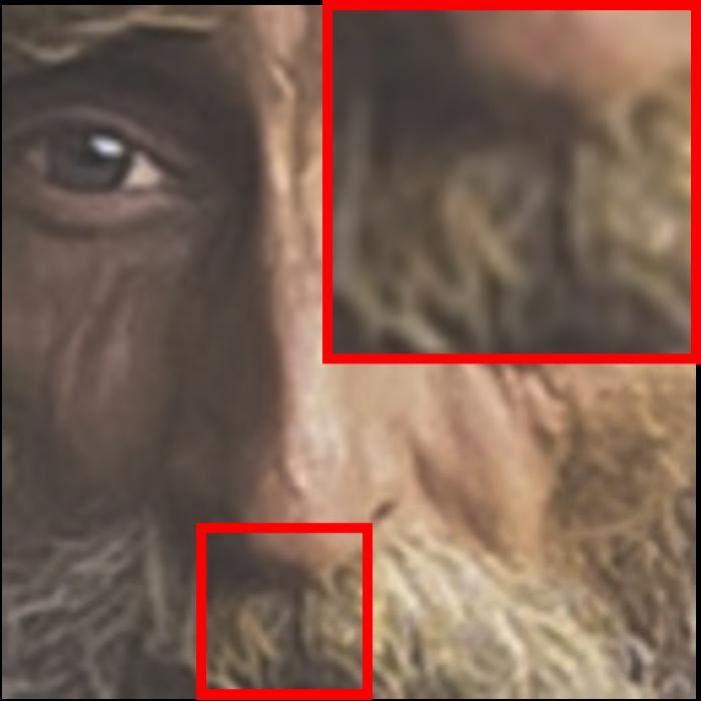}\\  
        \end{minipage}
        \hfill
        \begin{minipage}[h]{0.115\linewidth}
            \centering
            \includegraphics[width=1\linewidth]{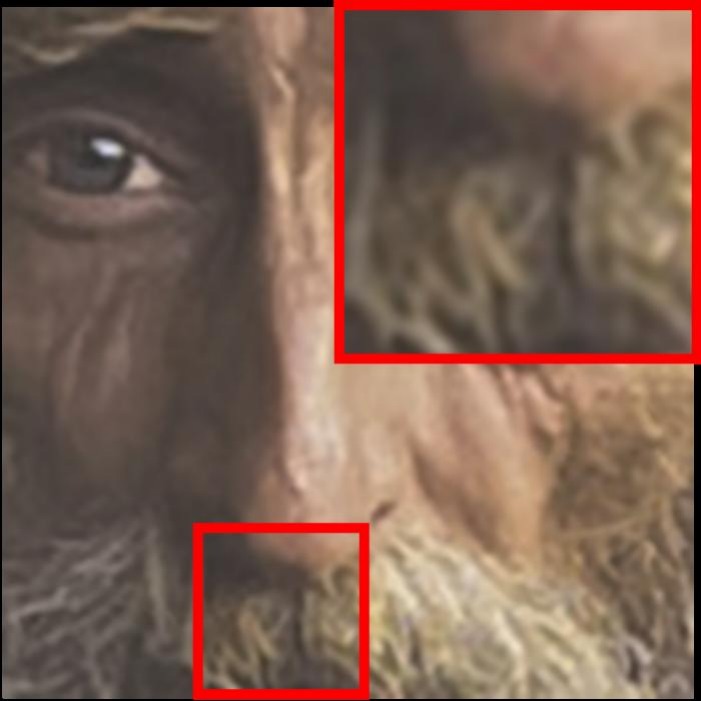}\\   
        \end{minipage}
        \hfill
        \begin{minipage}[h]{0.115\linewidth}
            \centering
            \includegraphics[width=1\linewidth]{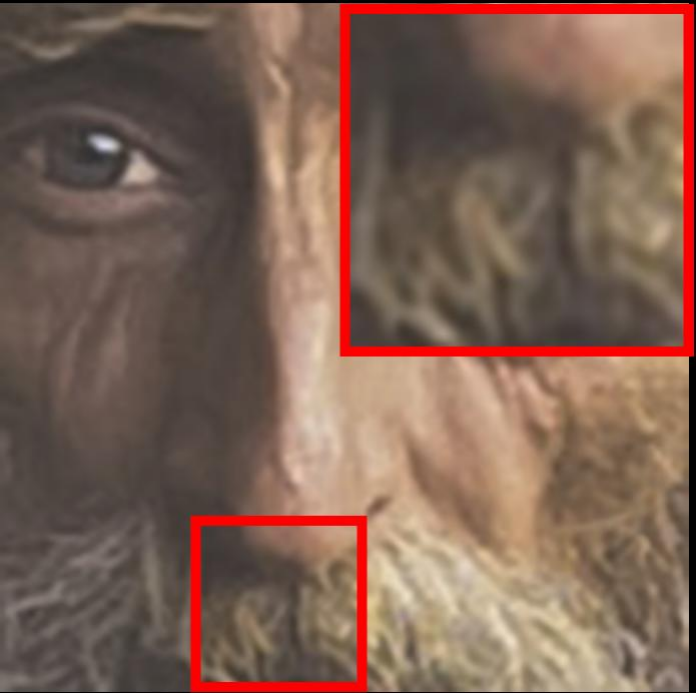}\\
        \end{minipage}
        \hfill
        \begin{minipage}[h]{0.115\linewidth}
            \centering
            \includegraphics[width=1\linewidth]{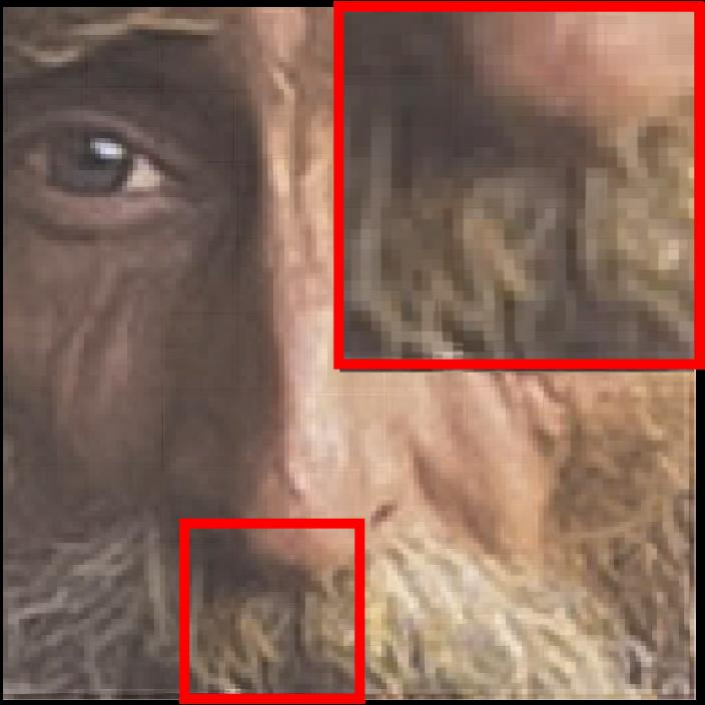}\\
        \end{minipage}
        \hfill
        \begin{minipage}[h]{0.115\linewidth}
            \centering
            \includegraphics[width=1\linewidth]{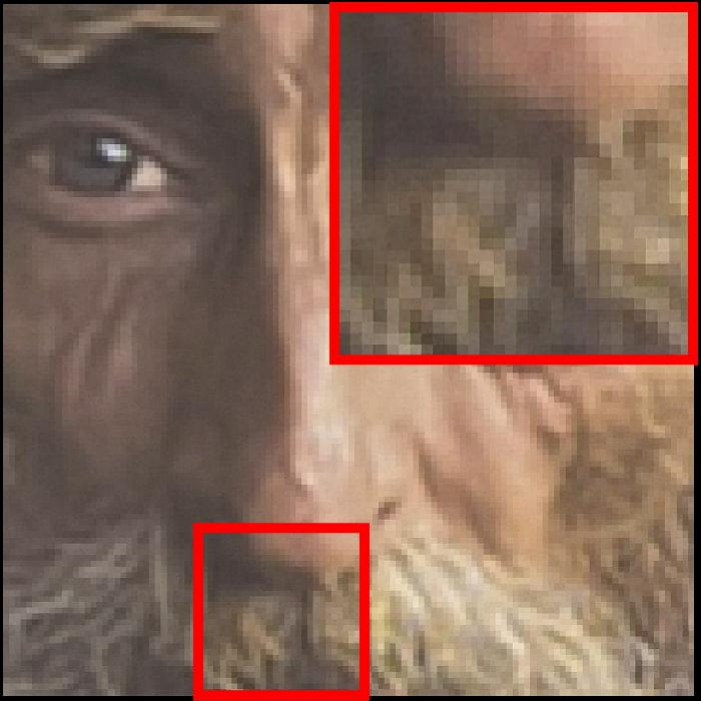}\\
        \end{minipage}
        \hfill
        \begin{minipage}[h]{0.115\linewidth}
            \centering
            \includegraphics[width=1\linewidth]{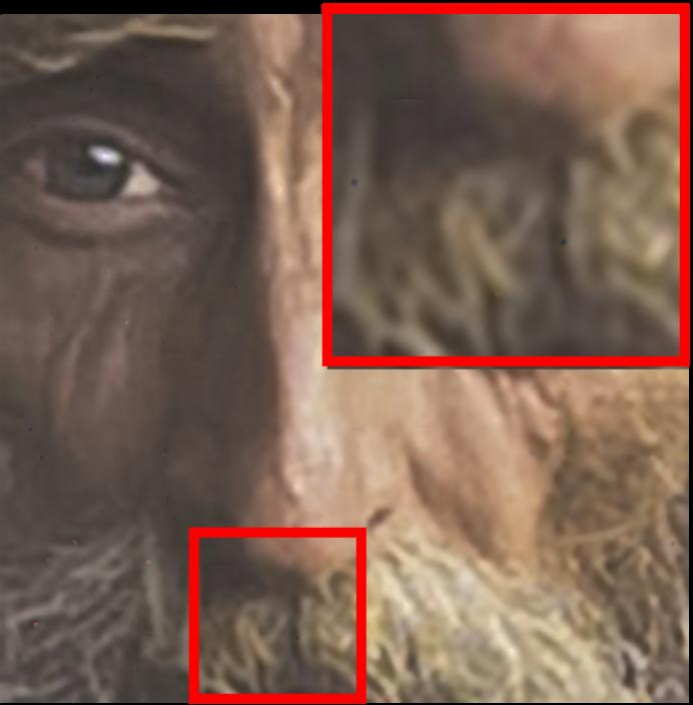}\\
        \end{minipage}
        \hfill
        \begin{minipage}[h]{0.115\linewidth}
            \centering
            \includegraphics[width=1\linewidth]{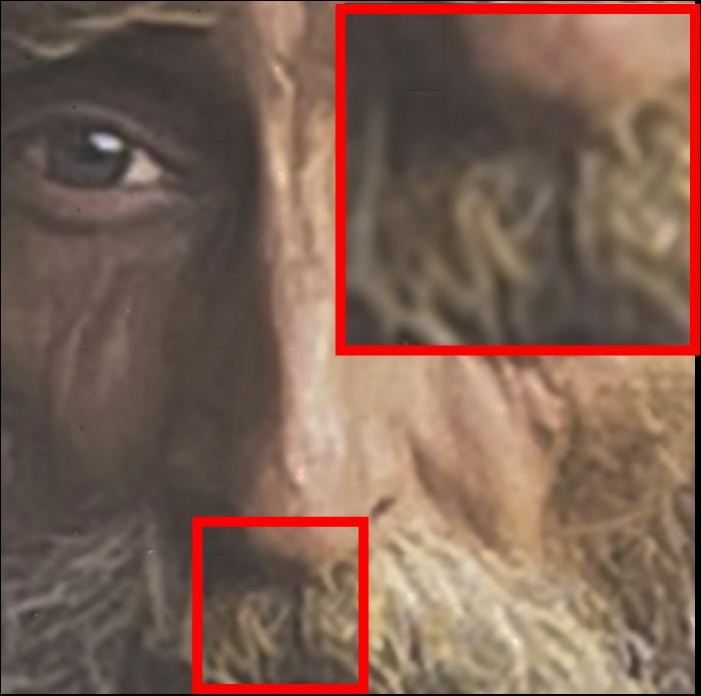}\\
        \end{minipage}
        \hfill
        \begin{minipage}[h]{0.115\linewidth}
            \centering
            \includegraphics[width=1\linewidth]{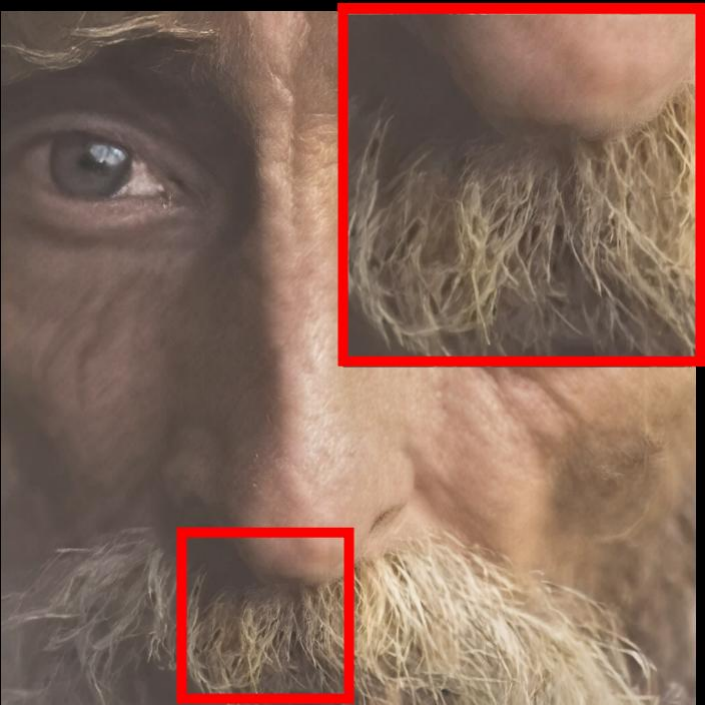}\\
        \end{minipage}
    \end{minipage}
    \vspace{0.01\linewidth}
    \begin{minipage}[h]{1\linewidth}
        \centering
        \begin{minipage}[h]{0.115\linewidth}
            \centering
            \includegraphics[width=1\linewidth]{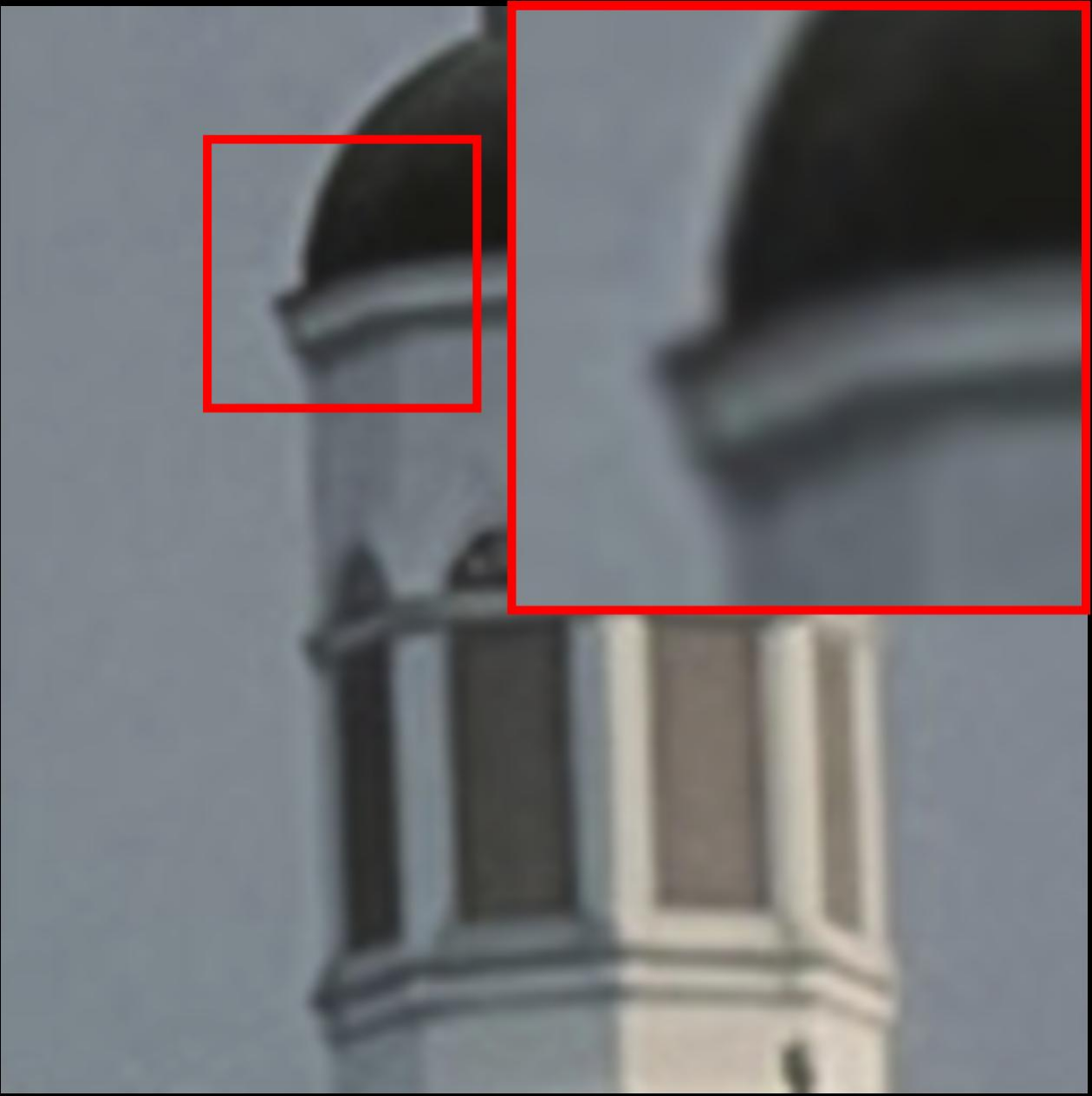}\\  
            \scriptsize{Bicubic}
        \end{minipage}
        \hfill
        \begin{minipage}[h]{0.115\linewidth}
            \centering
            \includegraphics[width=1\linewidth]{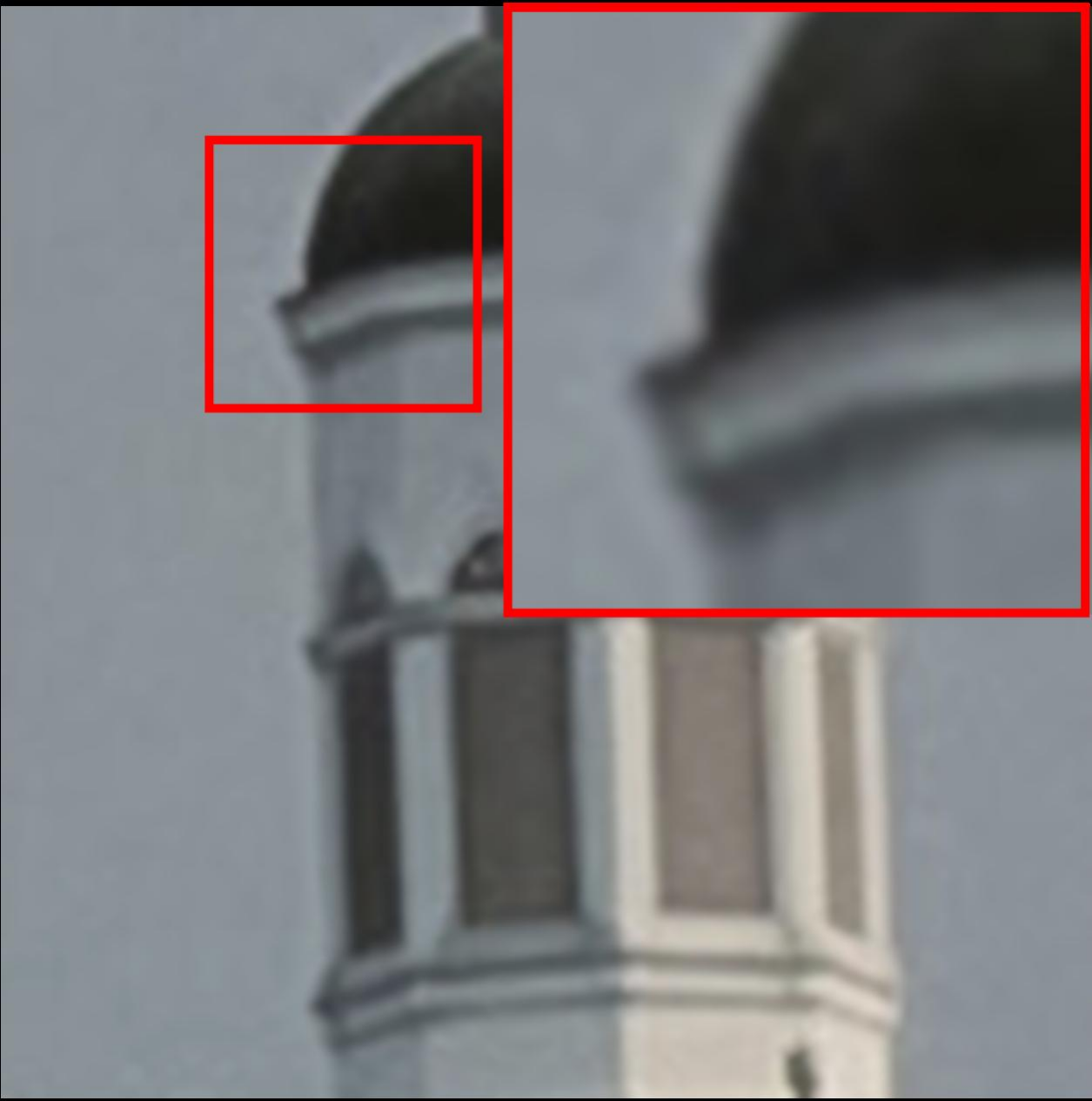}\\   
            \scriptsize{LINF}
        \end{minipage}
        \hfill
        \begin{minipage}[h]{0.115\linewidth}
            \centering
            \includegraphics[width=1\linewidth]{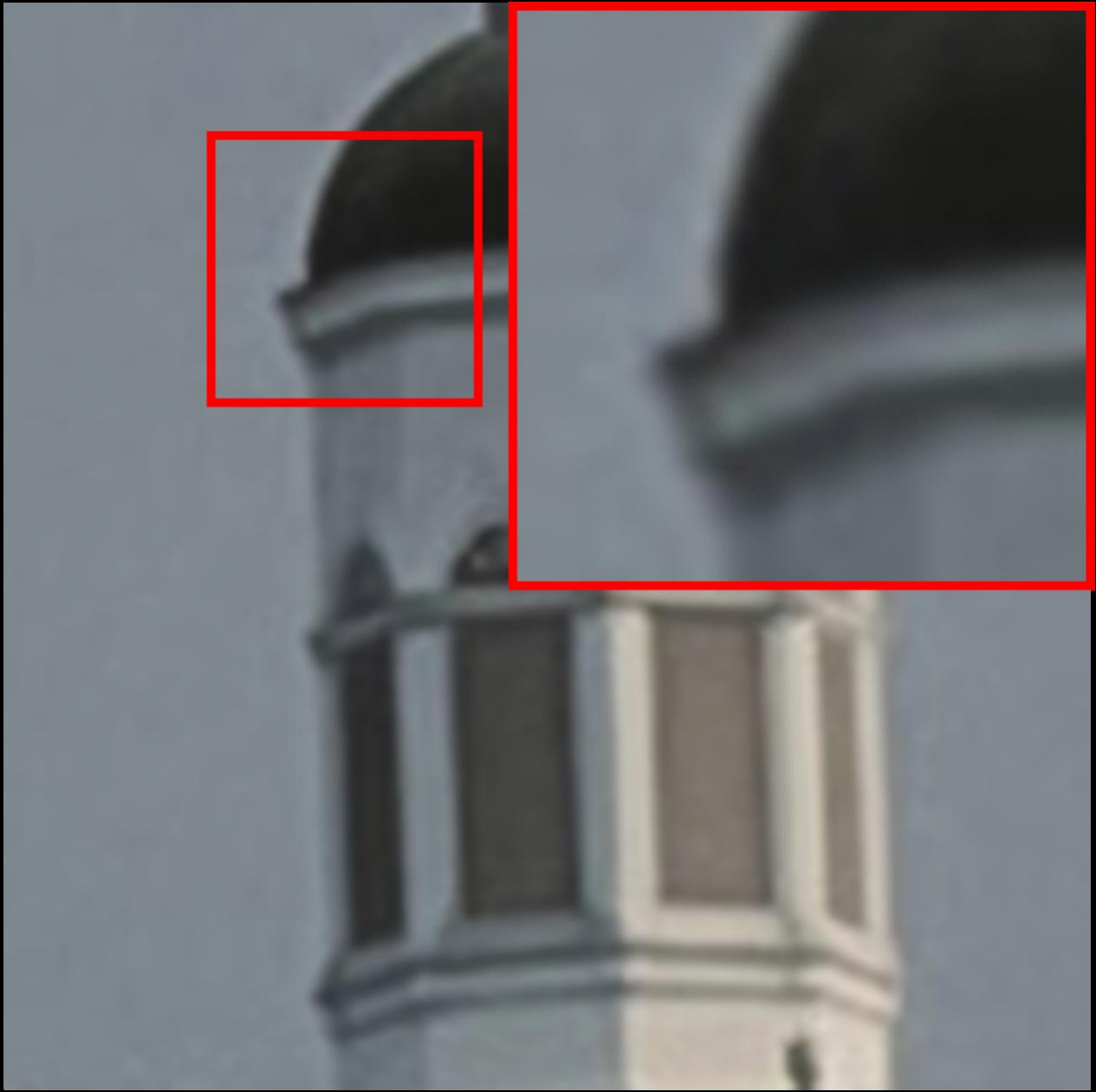}\\
            \scriptsize{BFSR}
        \end{minipage}
        \hfill
        \begin{minipage}[h]{0.115\linewidth}
            \centering
            \includegraphics[width=1\linewidth]{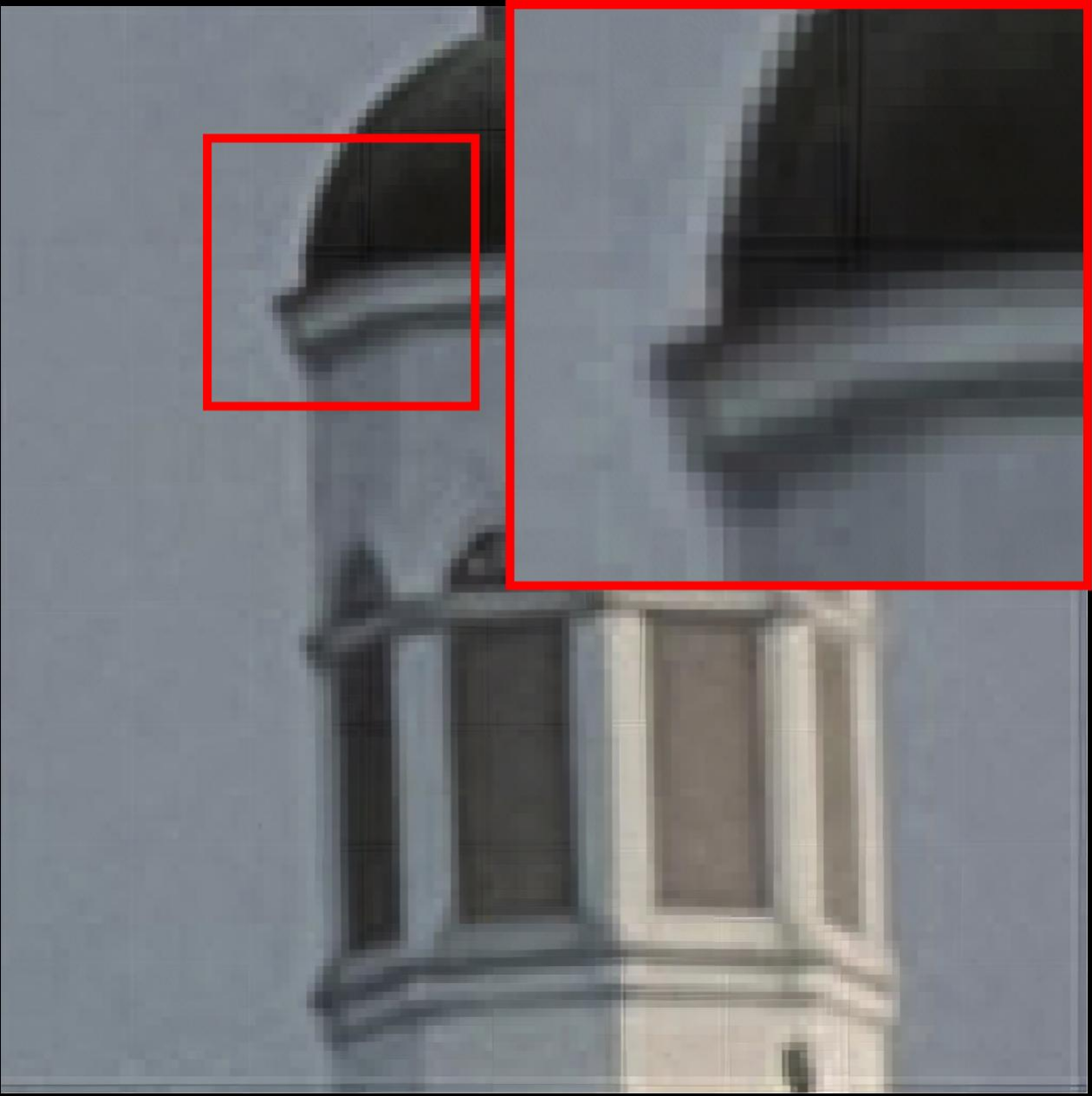}\\
            \scriptsize{IDM}
        \end{minipage}
        \hfill
        \begin{minipage}[h]{0.115\linewidth}
            \centering
            \includegraphics[width=1\linewidth]{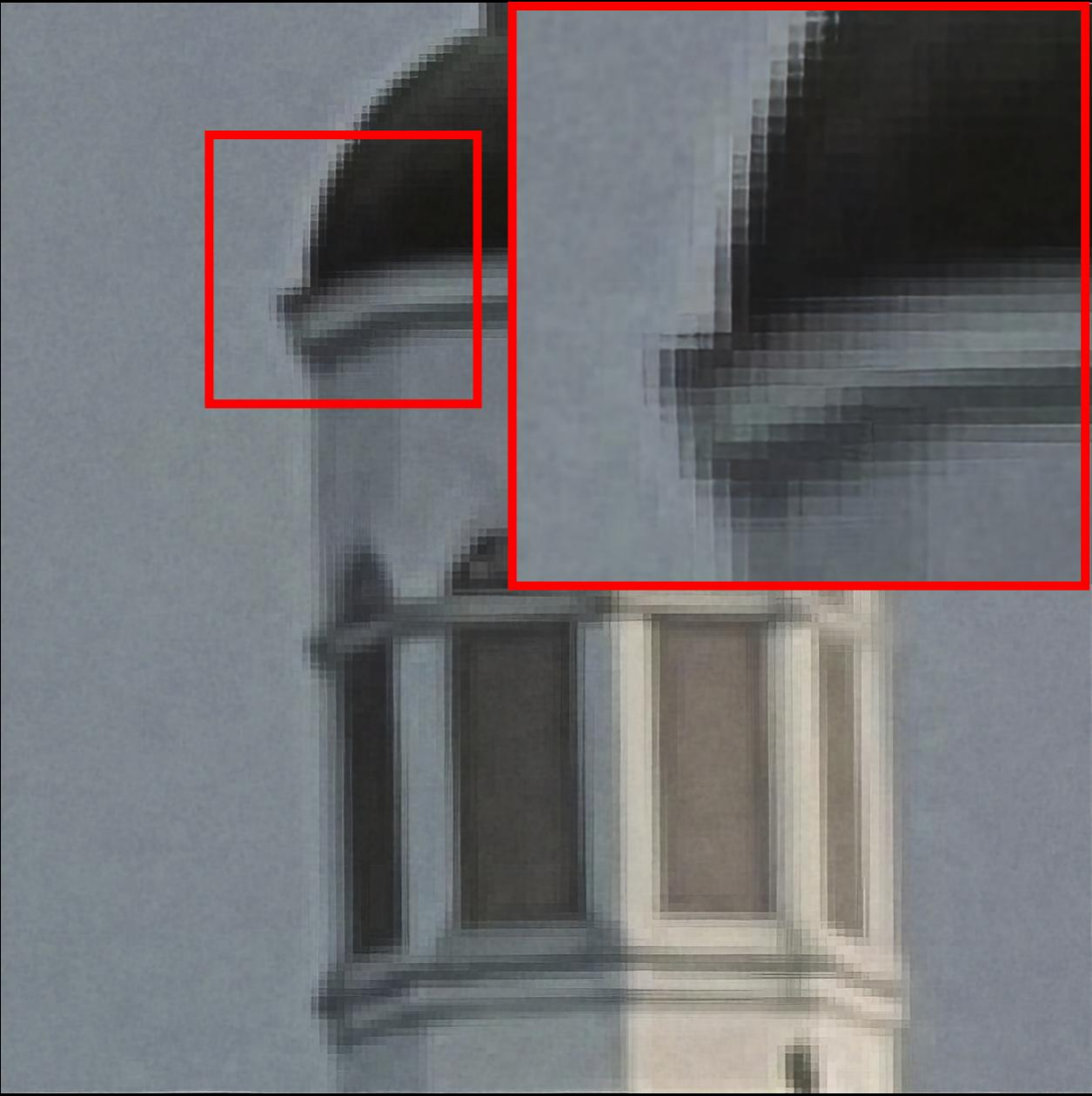}\\
            \scriptsize{Kim}
        \end{minipage}
        \hfill
        \begin{minipage}[h]{0.115\linewidth}
            \centering
            \includegraphics[width=1\linewidth]{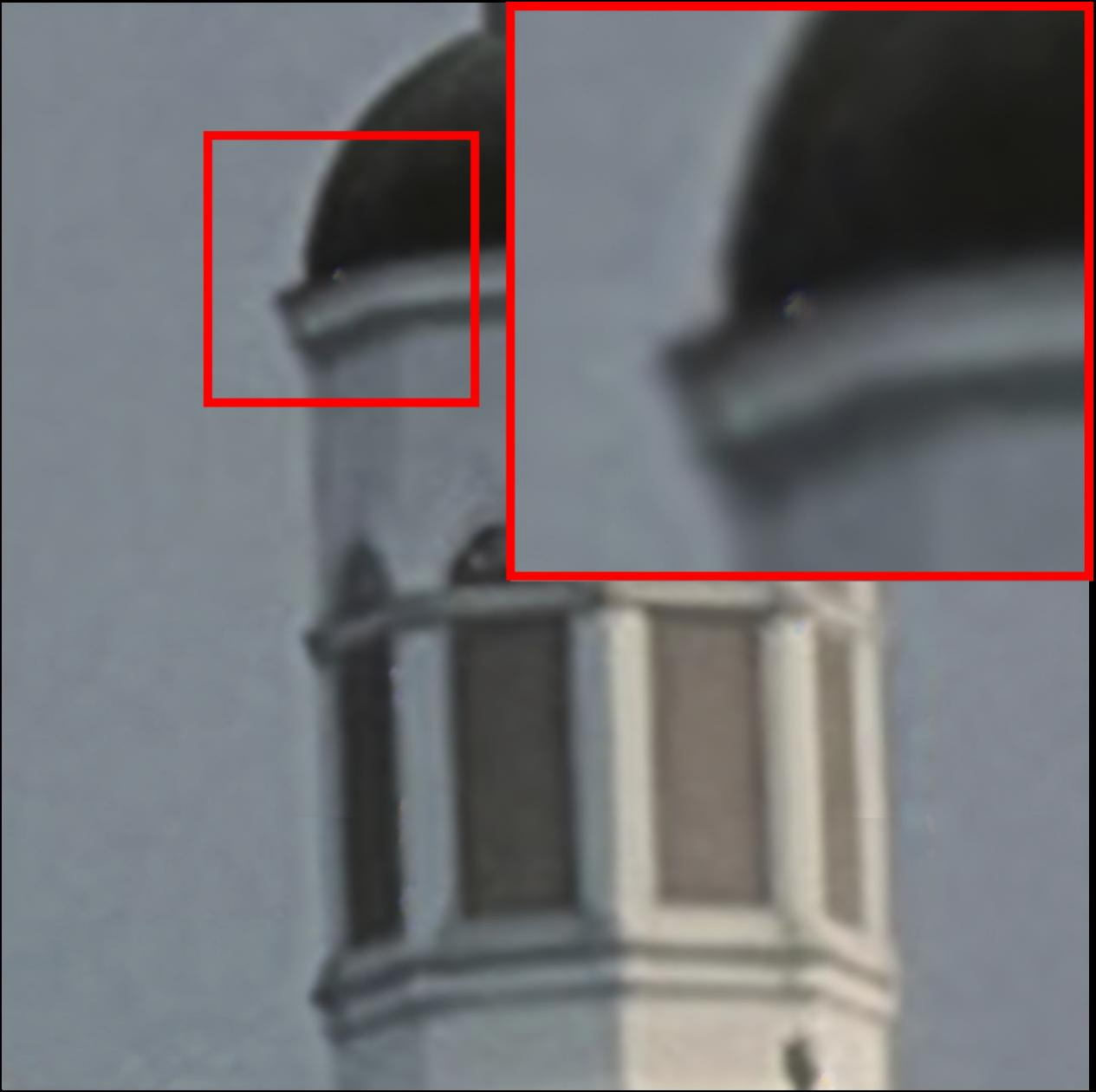}\\
            \scriptsize{LIIF+Diff}
        \end{minipage}
        \hfill
        \begin{minipage}[h]{0.115\linewidth}
            \centering
            \includegraphics[width=1\linewidth]{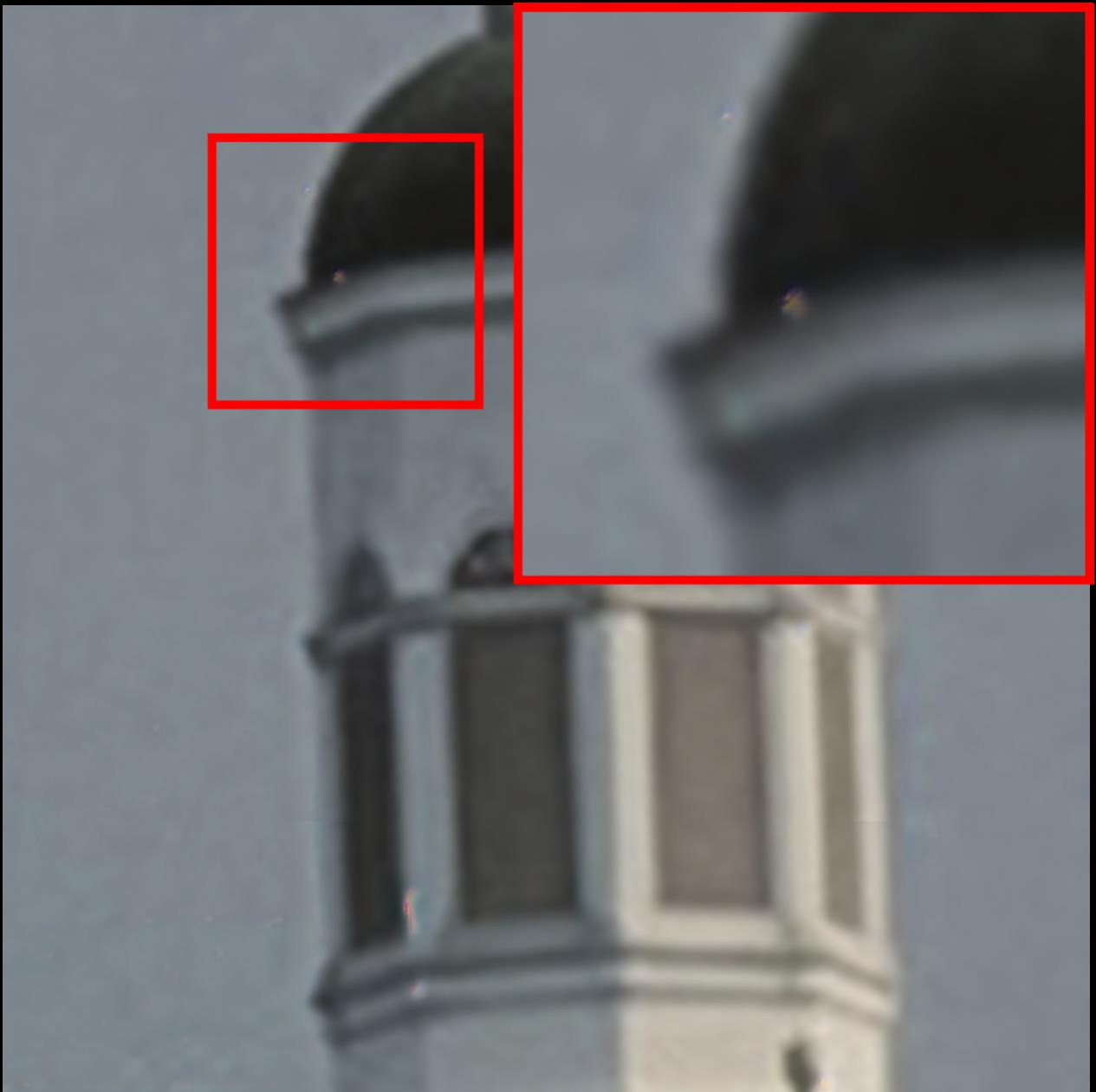}\\
            \scriptsize{CiaoSR+Diff}
        \end{minipage}
        \hfill
        \begin{minipage}[h]{0.115\linewidth}
            \centering
            \includegraphics[width=1\linewidth]{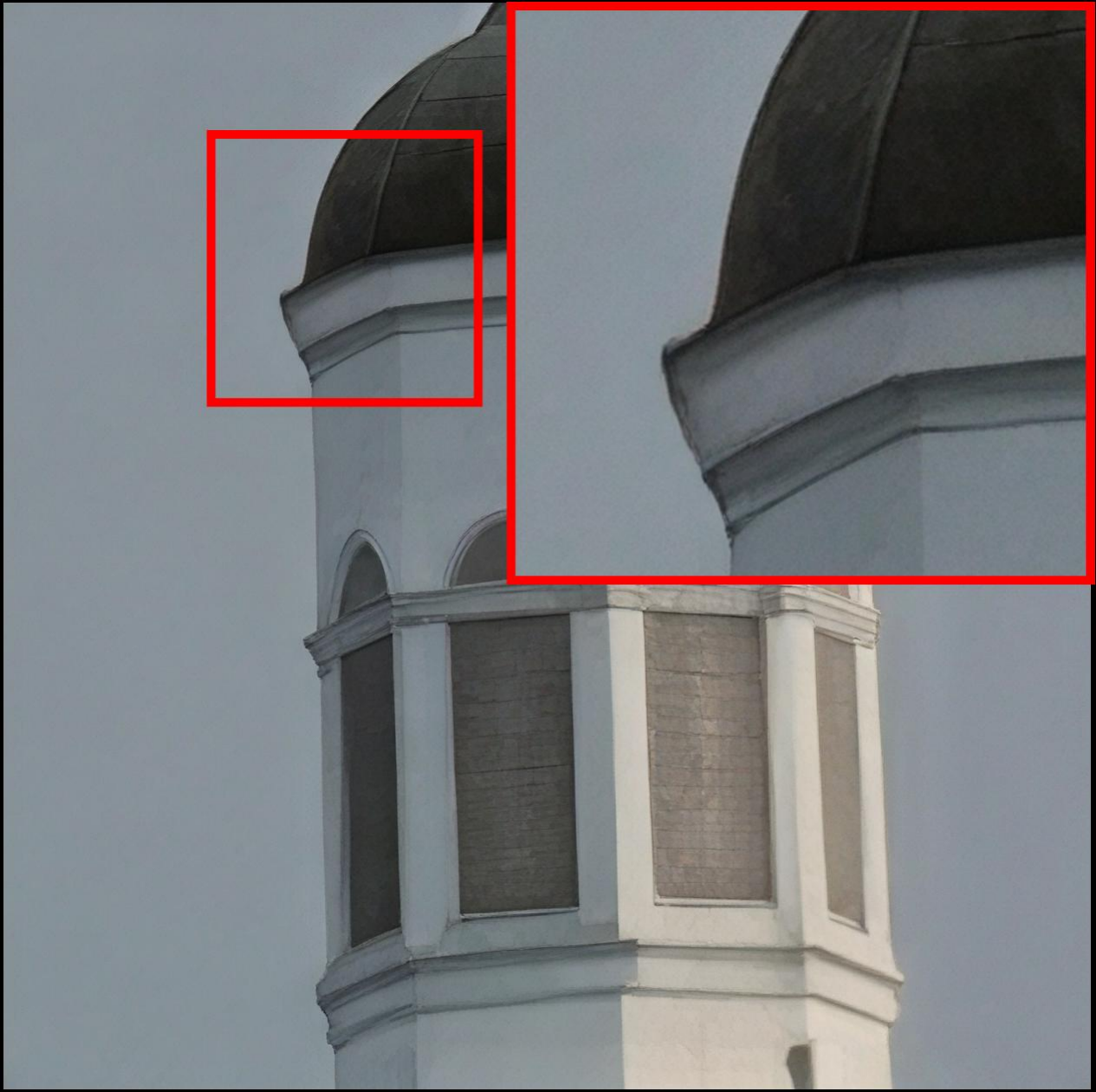}\\
            \scriptsize{CASR}
        \end{minipage}
    \end{minipage}
\end{minipage}

\centering
\caption{Qualitative comparison with different methods on the RealSR dataset. Our method produces clearer and more natural results.}\label{fig:real}
\vspace{-1em}
\end{figure*}

\subsubsection{Experiments on Synthetic Dataset}
For the synthetic DIV8K dataset,  the table \ref{table:div8K-large} presents a detailed comparison with baseline methods. Our method consistently achieves the highest perceptual quality across all upsampling factors. Notably, at $\times 30$, it outperforms the second-best method, LIIF+Diff, by 16.9\% in LPIPS. Regarding no-reference metrics (MUSIQ, NIQE, PI), our approach surpasses IDM by 75.2\%, 12.3\%, and 15.8\%, respectively. Importantly, performance remains robust even at extreme upsampling scales, while BFSR and other baselines exhibit significant degradation.

Qualitative comparisons in Fig.~\ref{fig:div8k} further corroborate these quantitative findings. At extreme magnifications, LINF and BFSR produce excessively blurry results, while IDM and Kim suffer from severe blocky artifacts. Even \emph{LIIF+Diff} and \emph{CiaoSR+Diff} exhibit noticeable artifacts and unrealistic textures when operating beyond their training distribution. In contrast, CASR preserves sharp edges and intricate fine details, clearly demonstrating its superiority in perceptual quality.

To investigate the impact of distribution shift on cyclic super-resolution, we evaluate all baseline methods under progressive upsampling ($\times4 \times 3 \times 1.5$) in the bottom-right column of Table~\ref{table:div8K-large}. Despite each upsampling stage operating within the training range, the overall performance does not improve compared to direct $\times18$ upsampling. The performance decline in the multi-stage approach is mainly due to error accumulation caused by distribution shift. Without effectively mitigating distribution shift, baseline methods struggle to benefit from cyclic cascading structures.

\subsubsection{Experiments on Real-World Datasets}
Table~\ref{table:real-world-large} presents the evaluation results on real-world datasets using no-reference metrics. Our approach consistently achieves superior perceptual quality. On the RealSR dataset at the $\times30$ scale, it outperforms the second-best method, IDM, by 34.1\%, 6.5\%, and 9.5\% in MUSIQ, NIQE, and PI, respectively. As shown in Fig.~\ref{fig:real}, baselines produce blocky artifacts or blurry edges, while CASR faithfully reconstructs fine textures and structural details.

\subsubsection{Face Comparisons with Diffusion-Based ASISR}
\begin{figure}[t]
\centering
\begin{minipage}[h]{1\linewidth}
    \begin{minipage}[h]{1\linewidth}
        \centering
        \begin{minipage}[h]{0.235\linewidth}
            \centering
            \scriptsize{128$\times$128}\\
            \includegraphics[width=1\linewidth]{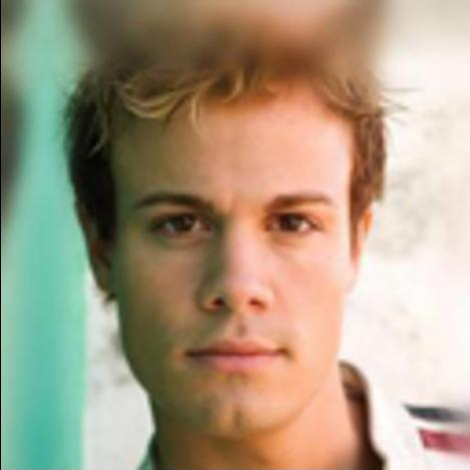}\\    
        \end{minipage}
        \hfill
        \begin{minipage}[h]{0.235\linewidth}
            \centering
            \scriptsize{1536$\times$1536}\\
            \includegraphics[width=1\linewidth]{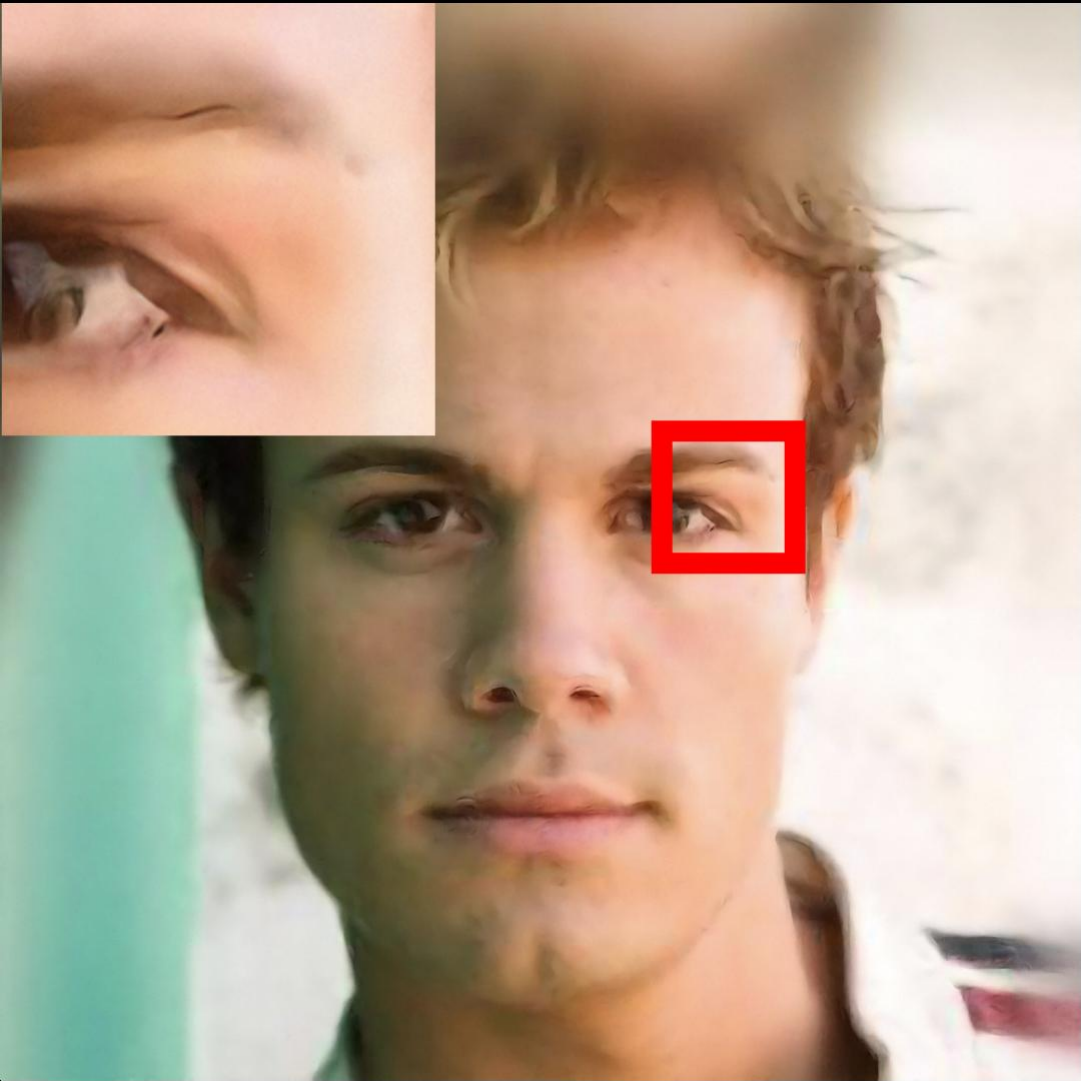}\\
        \end{minipage}
        \hfill
        \begin{minipage}[h]{0.235\linewidth}
            \centering
            \scriptsize{1536$\times$1536}\\
            \includegraphics[width=1\linewidth]{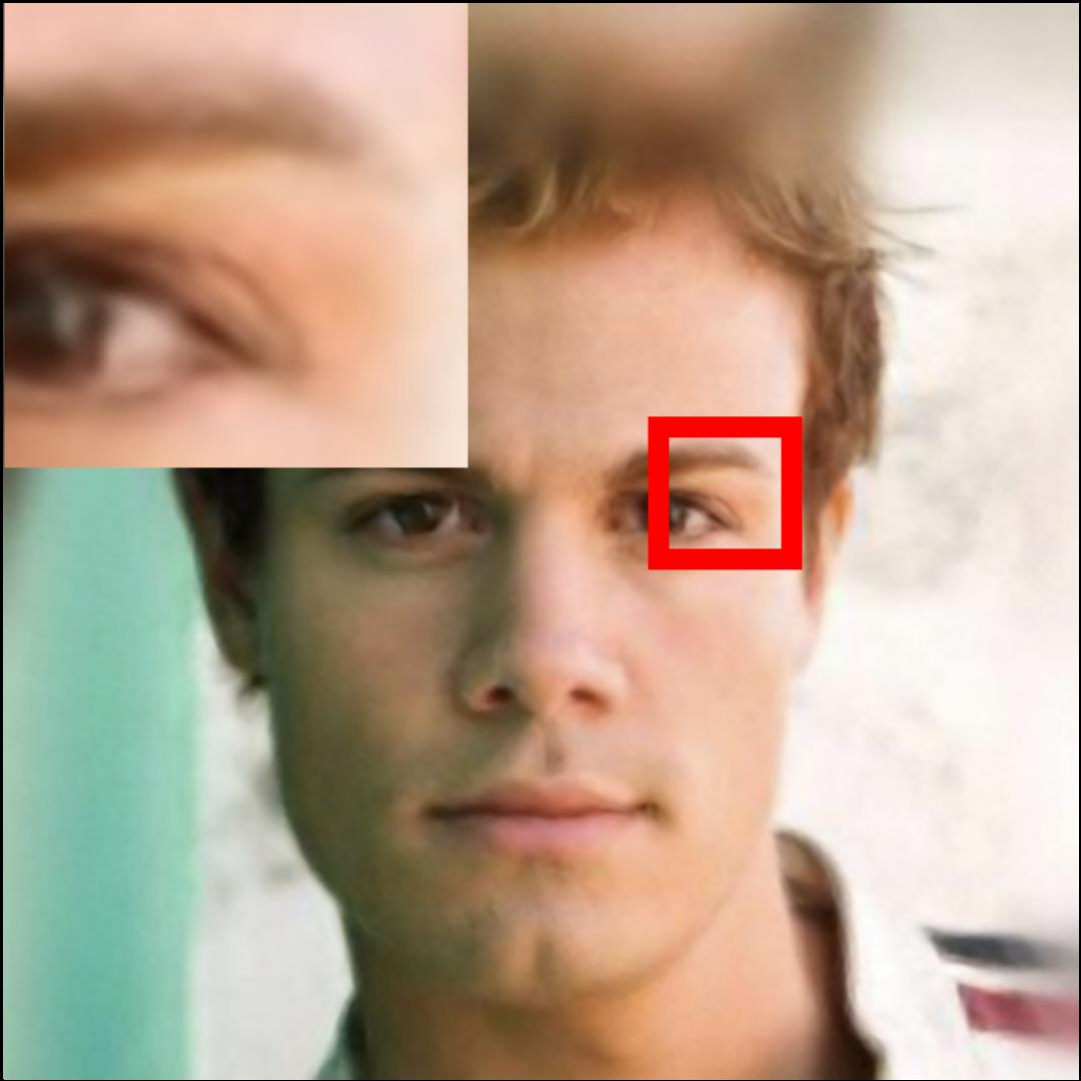}\\
        \end{minipage}
        \hfill
        \begin{minipage}[h]{0.235\linewidth}
            \centering
            \scriptsize{1536$\times$1536}\\
            \includegraphics[width=1\linewidth]{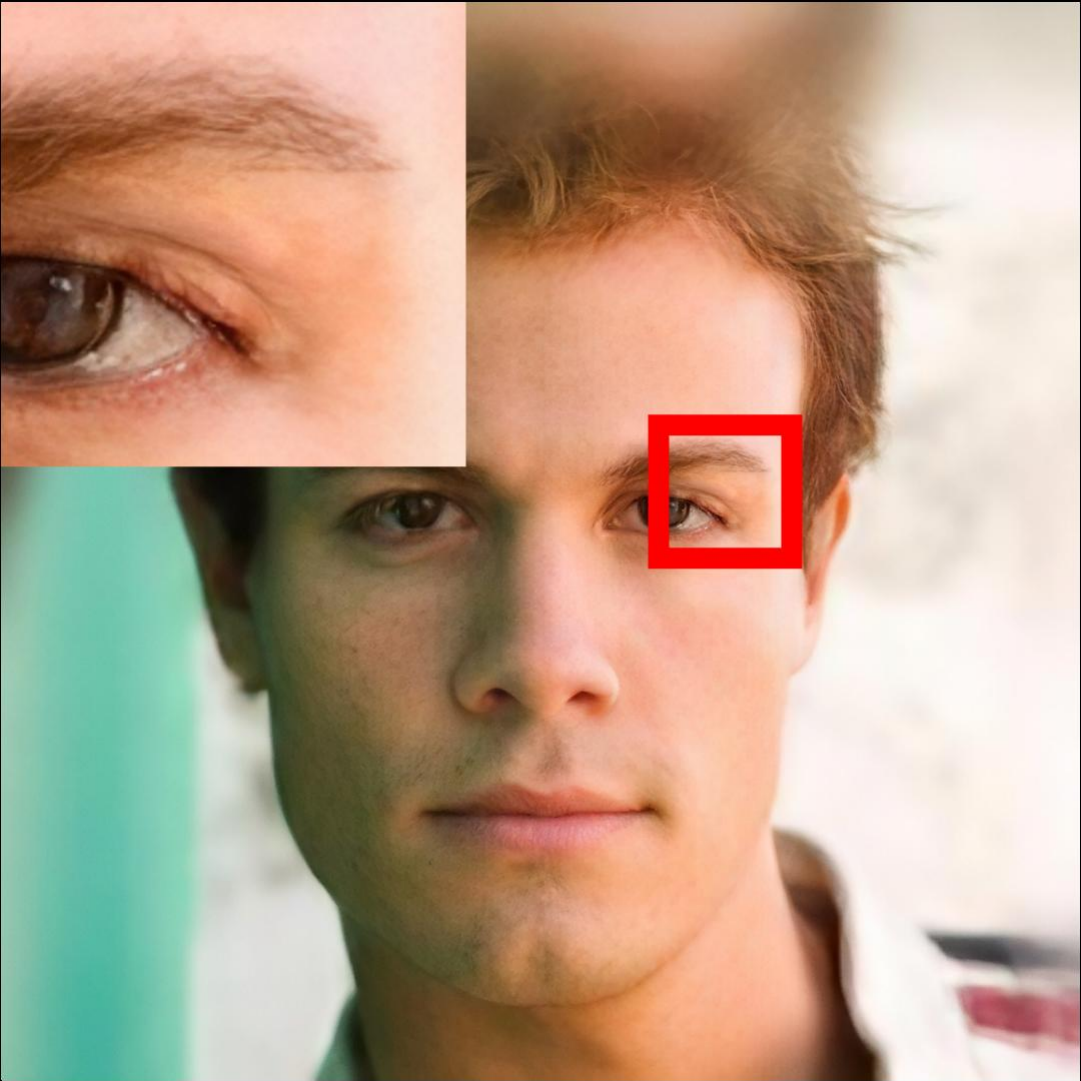}
        \end{minipage}
    \end{minipage}
    \vspace{0.001\linewidth}\\
    \begin{minipage}[h]{1\linewidth}
        \centering
        \begin{minipage}[h]{0.235\linewidth}
            \centering
            \includegraphics[width=1\linewidth]{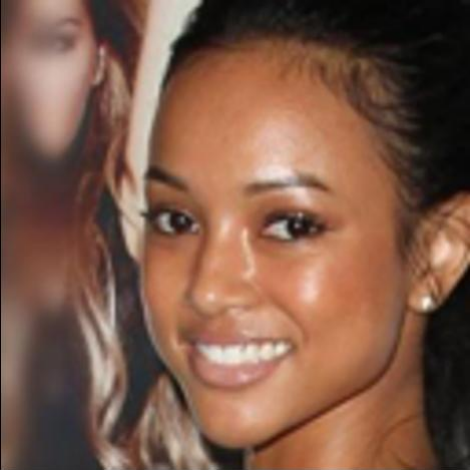}\\
            \scriptsize{LR}
        \end{minipage}
        \hfill
        \begin{minipage}[h]{0.235\linewidth}
            \centering
            \includegraphics[width=1\linewidth]{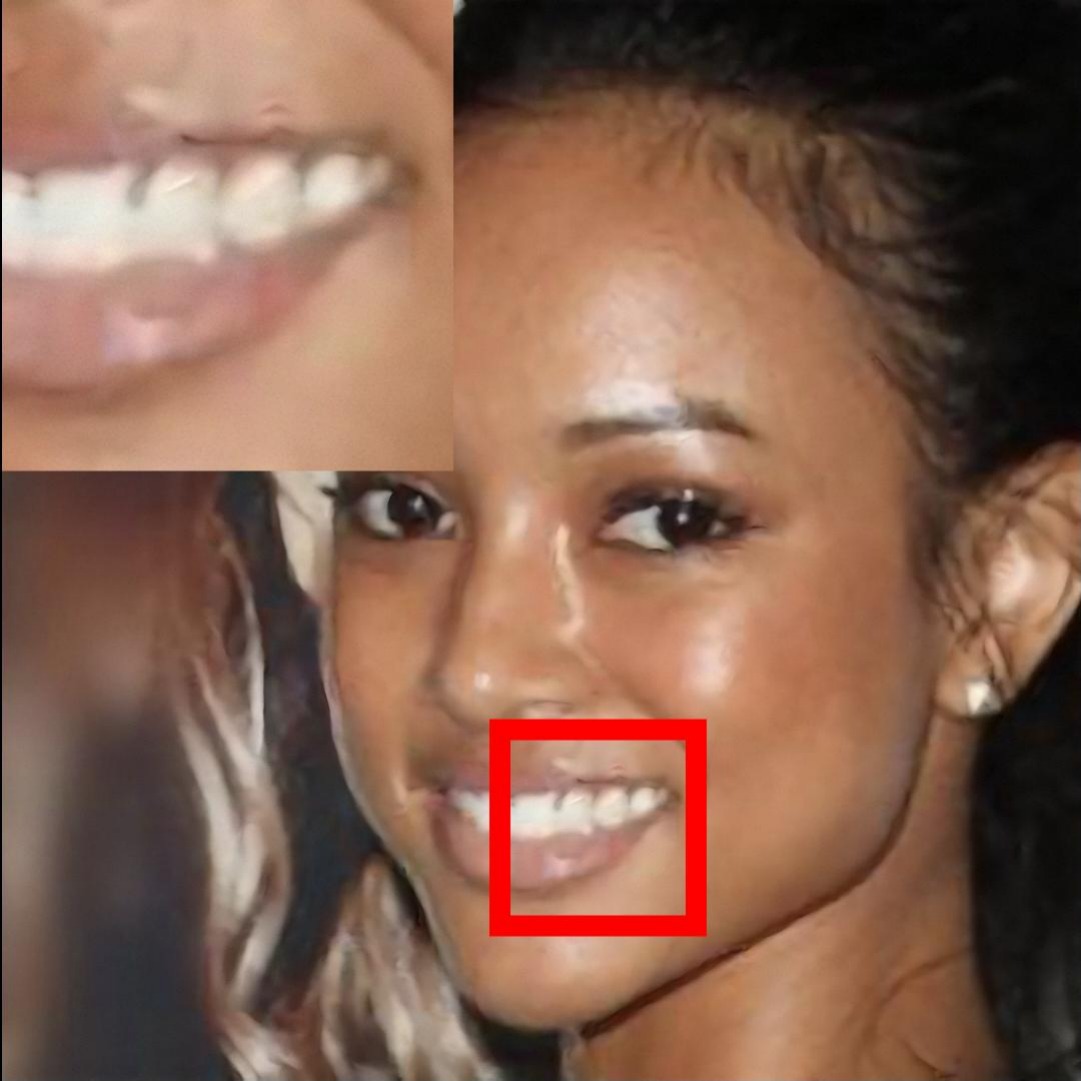}\\
            \scriptsize{\textbf{IDM}}
        \end{minipage}
        \hfill
        \begin{minipage}[h]{0.235\linewidth}
            \centering
            \includegraphics[width=1\linewidth]{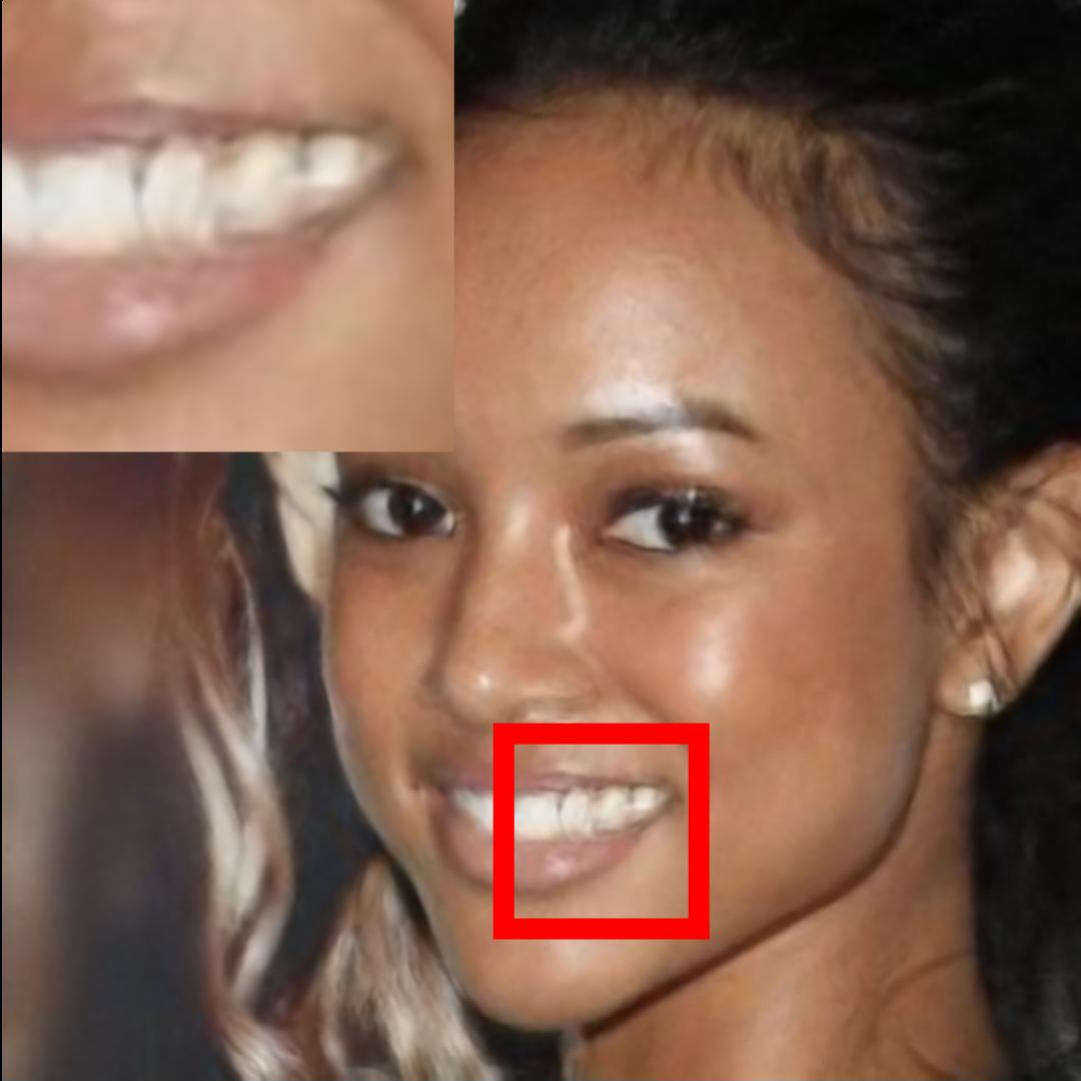}\\
            \scriptsize{\textbf{Kim}}
        \end{minipage}
        \hfill
        \begin{minipage}[h]{0.235\linewidth}
            \centering
            \includegraphics[width=1\linewidth]{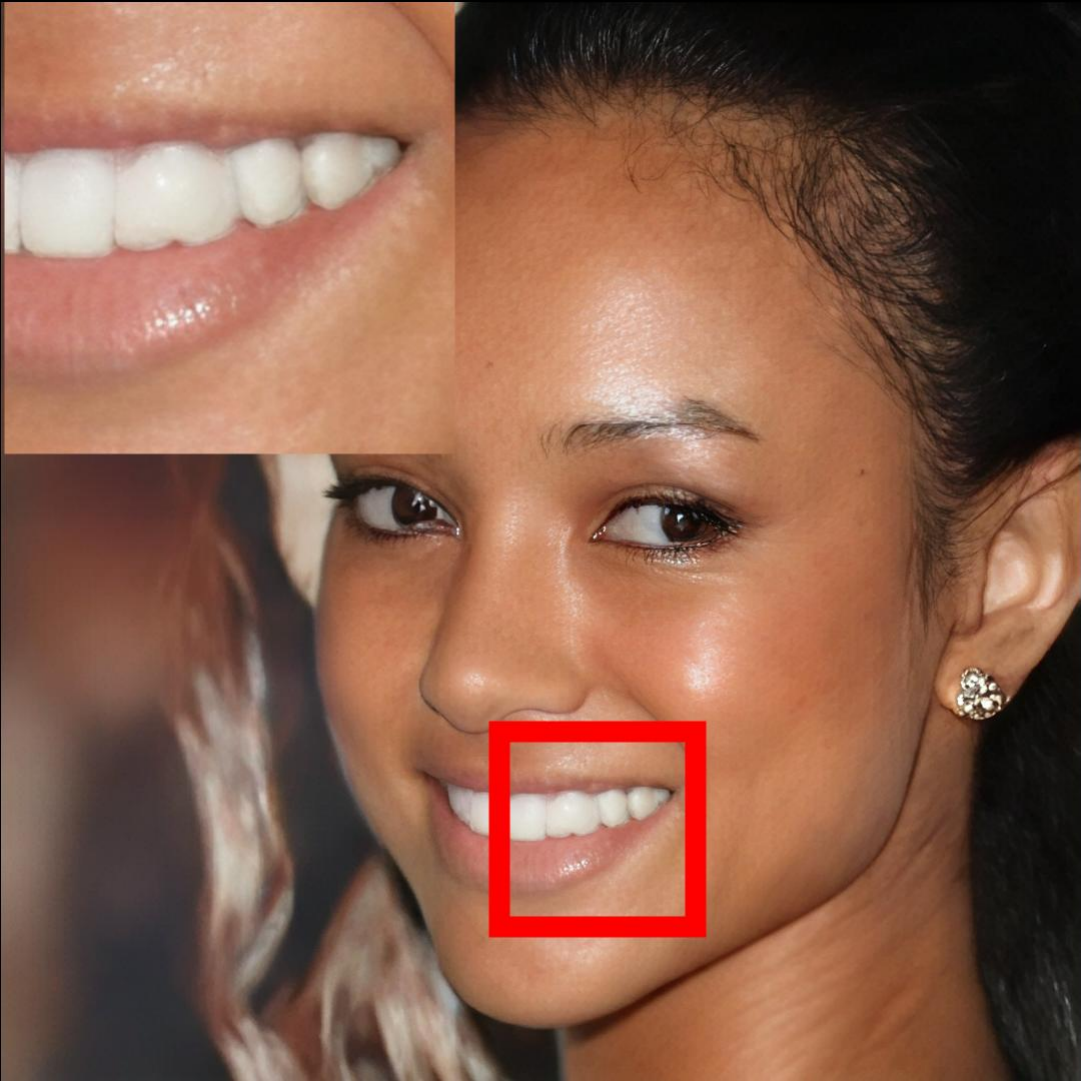}\\
            \scriptsize{\textbf{CASR}}
    \end{minipage}
    \end{minipage}
\end{minipage}

\centering
\caption{Super-resolution results at \( \times 12 \) on CelebA-HQ. IDM and Kim fail to recover fine facial details, while our method produces cleaner and sharper reconstructions.\label{fig:celeba-hq}}
\end{figure}

Table~\ref{table:celeb-hq} presents results on CelebA-HQ across various upsampling scales. While IDM achieves acceptable performance at lower scales, its quality deteriorates as the upsampling factor increases. Fig.~\ref{fig:celeba-hq} illustrates $\times 12$ results ($1536 \times 1536$). IDM and Kim produce overly smooth facial reconstructions, whereas CASR accurately restores fine facial features such as eyes and mouth.
\begin{table}[t]
    \scriptsize
    \centering
    \caption{Comparison with diffusion-based ASISR methods on the CelebA-HQ dataset. Our method achieves superior performance at large upsampling scales. \label{table:celeb-hq}} 
    \resizebox{0.45\textwidth}{!}{
    \begin{tabular}{  l | c c c | c c c }      
        \toprule
        \multirow{3}{*}{Method}  &  \multicolumn{6}{c}{\textbf{CelebA-HQ}}\\
        \multirow{2}{*}{} & \multicolumn{3}{c|}{$\times$4} & \multicolumn{3}{c}{$\times$6} \\

        \multirow{2}{*}{} & MUSIQ$\uparrow$ & NIQE$\downarrow$  &  PI$\downarrow$  & MUSIQ$\uparrow$  & NIQE$\downarrow$  &  PI$\downarrow$ \\

        \midrule
        IDM  \cite{idm} &	\textbf{72.79}&9.70&\textbf{6.97}&66.77&5.60&5.44\\
        Kim \cite{kim2024arbitrary} & 43.57&\textbf{7.83}&7.48&41.06&8.06&7.87\\
        CASR & 71.12&9.69&7.19&\textbf{70.86}&\textbf{4.77}&\textbf{4.87}\\
        \midrule

        \multirow{2}{*}{Method} & \multicolumn{3}{c|}{$\times$8} & \multicolumn{3}{c}{$\times$12} \\

        \multirow{2}{*}{} & MUSIQ$\uparrow$ & NIQE$\downarrow$  &  PI$\downarrow$  & MUSIQ$\uparrow$  & NIQE$\downarrow$  &  PI$\downarrow$ \\

        \midrule
         
         IDM \cite{idm} & 60.18&6.51&5.83&41.47&9.66&8.08\\
         Kim \cite{kim2024arbitrary} &39.40&9.15&8.21&31.66&11.10&9.56\\
         CASR &\textbf{72.41}&\textbf{4.63}&\textbf{4.17}&\textbf{71.71}&\textbf{4.77}&\textbf{4.04}\\
        \bottomrule

    \end{tabular}
    }
\end{table}

\subsection{Ablation Study}
\subsubsection{Component Effectiveness}
To evaluate the contribution of each component in the proposed \textbf{CASR} framework,
we conduct an ablation study with four model variants.
\textbf{Base Model}: A diffusion-based SR model built on SD-Turbo with LoRA fine-tuning, where LR inputs are upsampled via bicubic interpolation before HR reconstruction.
\textbf{Model 1}: Adds the \emph{superpixel segmentation} module.
\textbf{Model 2}: Further incorporates the \emph{depth-conditioning} module for geometric consistency.
\textbf{Full Model}: The complete version equipped with the SARM.

\begin{table}[!ht]
    \caption{Ablation study of major components in CASR. Each module contributes to the overall performance improvement.
    \label{table:abla_1}}
    \centering
    \scriptsize
    \resizebox{1\linewidth}{!}{
    \begin{tabular}{c c c |c c c c}
    \toprule
    \multirow{2}{*}{\textbf{SuperPixel}} & \multirow{2}{*}{\textbf{Depth}} & \multirow{2}{*}{\textbf{SARM}} & \multicolumn{4}{c}{\textbf{$\times$ 18} ($\times 4 \times 3 \times 1.5$)} \\
      {} & {} & {} & \bf{LPIPS$\downarrow$} & \bf{MUSIQ$\uparrow$} & \bf{NIQE$\downarrow$} & \bf{PI$\downarrow$} \\
    \midrule
     {\color{blue}\ding{55}}  & {\color{blue}\ding{55}} & {\color{blue}\ding{55}} &  0.585 & 31.73 & 7.10 & 5.91 \\
     {\color{red}\ding{51}}  & {\color{blue}\ding{55}} & {\color{blue}\ding{55}} &   0.471 & 42.23 & 6.61 & 5.96 \\
     {\color{red}\ding{51}}  & {\color{red}\ding{51}} & {\color{blue}\ding{55}} &  0.467 & 45.18 & 6.15 & 5.37 \\
     {\color{red}\ding{51}} & {\color{red}\ding{51}} & {\color{red}\ding{51}} &  0.450 & 51.44 & 6.01 & 5.24 \\
    \bottomrule
    \end{tabular}
    }
    \vspace{-1em}
\end{table}

\begin{figure}[!ht]
\centering
\begin{minipage}[h]{1\linewidth}
    \begin{minipage}[h]{1\linewidth}
        \centering
        \begin{minipage}[h]{0.153\linewidth}
            \centering
            \includegraphics[width=1\linewidth]{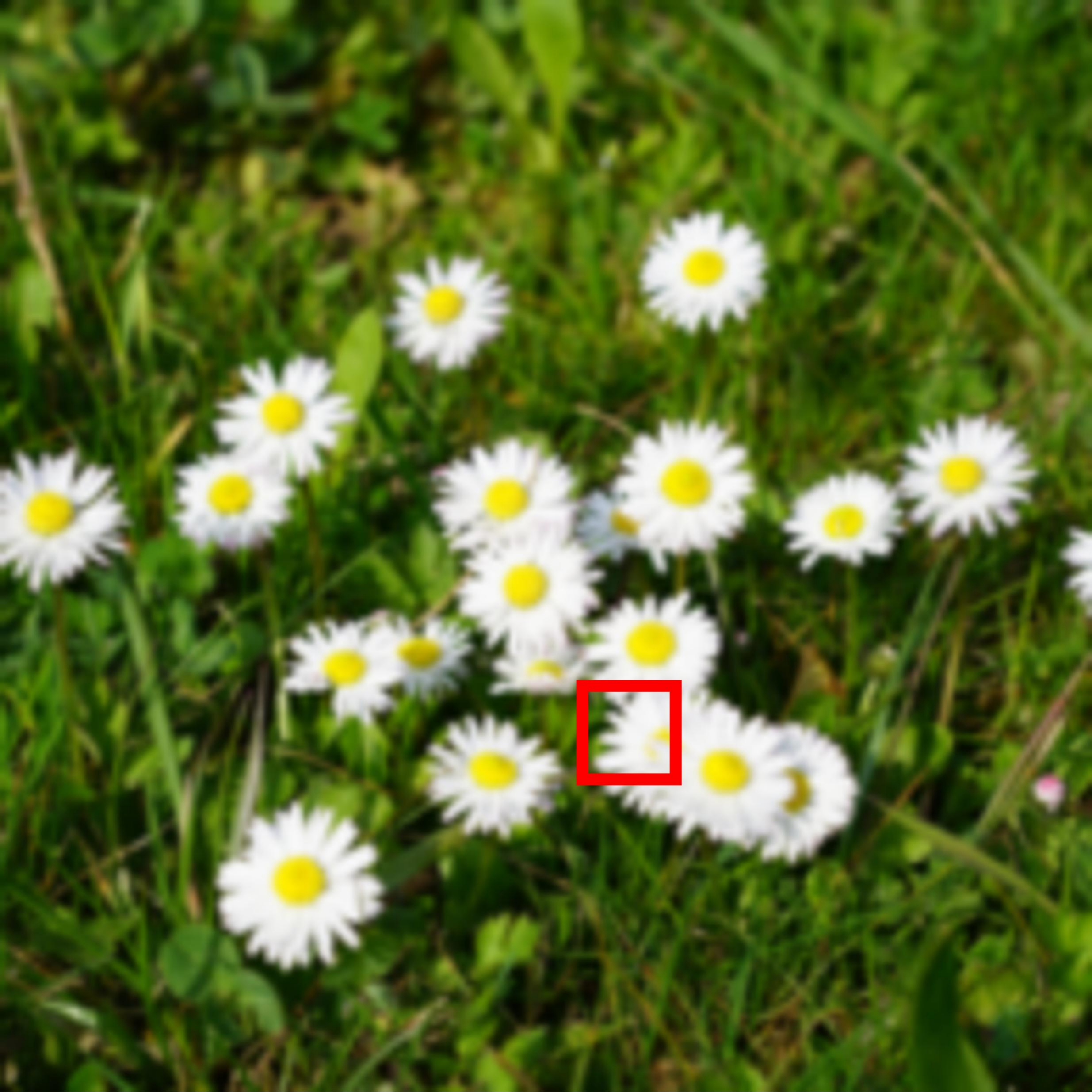}\\  
        \end{minipage}
        \hfill
        \begin{minipage}[h]{0.153\linewidth}
            \centering
            \includegraphics[width=1\linewidth]{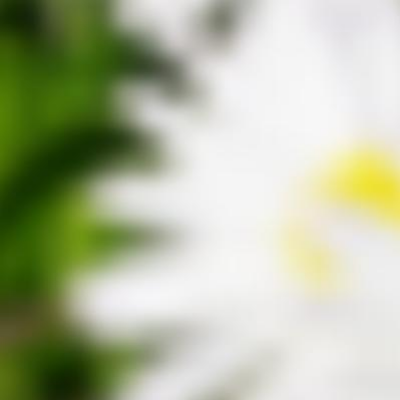}\\    
        \end{minipage}
        \hfill
        \begin{minipage}[h]{0.153\linewidth}
            \centering
            \includegraphics[width=1\linewidth]{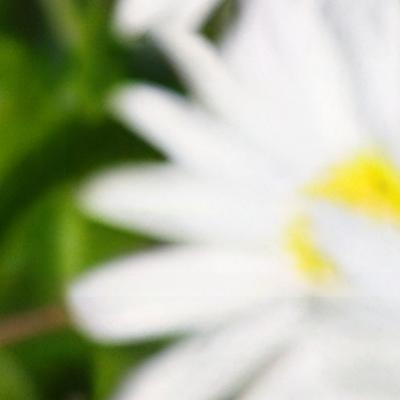}\\
        \end{minipage}
        \hfill
        \begin{minipage}[h]{0.153\linewidth}
            \centering
            \includegraphics[width=1\linewidth]{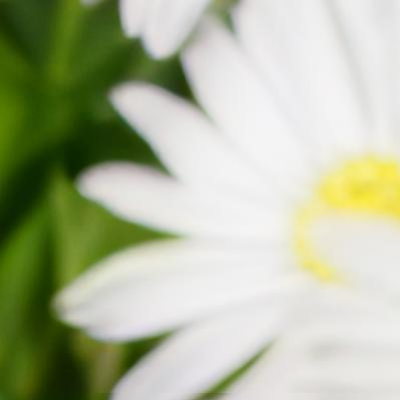}\\
        \end{minipage}
        \hfill
        \begin{minipage}[h]{0.153\linewidth}
            \centering
            \includegraphics[width=1\linewidth]{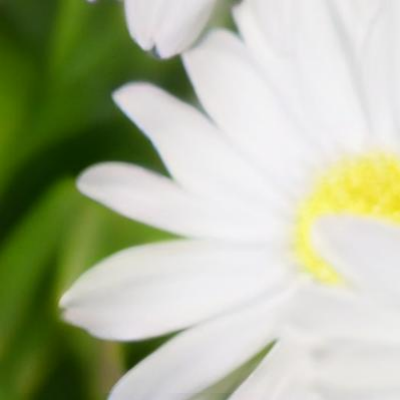}
        \end{minipage}
        \hfill
        \begin{minipage}[h]{0.153\linewidth}
            \centering
            \includegraphics[width=1\linewidth]{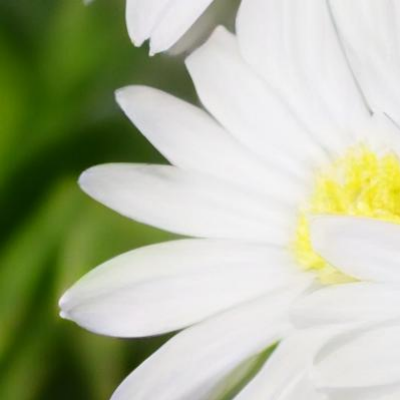}
        \end{minipage}
    \end{minipage}
    \vspace{0.005\linewidth}\\
    \begin{minipage}[h]{1\linewidth}
        \centering
        \begin{minipage}[h]{0.153\linewidth}
            \centering
            \includegraphics[width=1\linewidth]{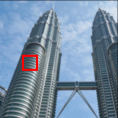}\\  
            \scriptsize{Origin}
        \end{minipage}
        \hfill
        \begin{minipage}[h]{0.153\linewidth}
            \centering
            \includegraphics[width=1\linewidth]{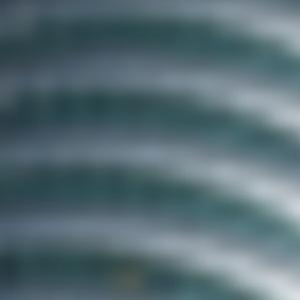}\\    
            \scriptsize{LR}
        \end{minipage}
        \hfill
        \begin{minipage}[h]{0.153\linewidth}
            \centering
            \includegraphics[width=1\linewidth]{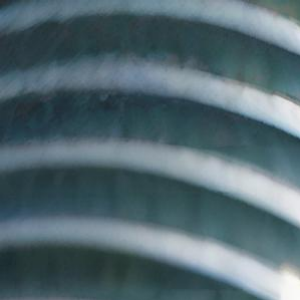}\\
            \scriptsize{Base}
        \end{minipage}
        \hfill
        \begin{minipage}[h]{0.153\linewidth}
            \centering
            \includegraphics[width=1\linewidth]{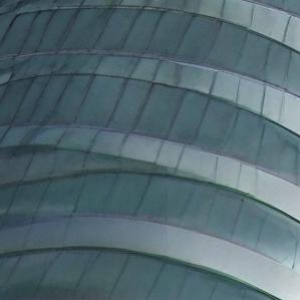}\\
            \scriptsize{Model1}
        \end{minipage}
        \hfill
        \begin{minipage}[h]{0.153\linewidth}
            \centering
            \includegraphics[width=1\linewidth]{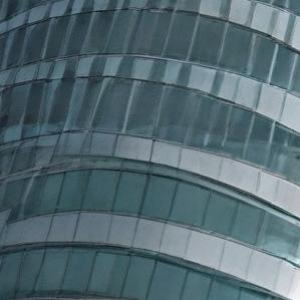}
            \scriptsize{Model2}
        \end{minipage}
        \hfill
        \begin{minipage}[h]{0.153\linewidth}
            \centering
            \includegraphics[width=1\linewidth]{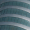}
            \scriptsize{Full Model}
        \end{minipage}
    \end{minipage}
\end{minipage}

\centering
\caption{ Impact of different components. Incorporating the superpixel module (Model1) effectively suppresses accumulated blur and artifacts during cascading, while depth conditioning (Model2) further enhances edge sharpness. The full model produces natural results with consistent textures across patches.
}\label{fig:ablation_1}
\end{figure}

Quantitative results in Table~\ref{table:abla_1} and qualitative comparisons in Fig.~\ref{fig:ablation_1} collectively validate the effectiveness of each component.
Removing any module consistently degrades perceptual quality, confirming their complementary roles.
The \textbf{Base Model} suffers from distribution shifts during iterative refinement, producing blurred edges and artifacts in uniform regions (e.g., building facades).
Introducing the superpixel segmentation module (\textbf{Model 1}) sharpens boundaries and restores fine textures, indicating that spatially coherent segmentation stabilizes feature distributions across iterations.
Adding the depth-conditioning module (\textbf{Model 2}) further enhances geometric fidelity---petal structures become clearer and depth transitions smoother---though patch-level inconsistencies remain in repetitive patterns such as windows.
The \textbf{Full Model} incorporates the SARM, yielding substantial gains across all metrics, demonstrating that self-similarity-aware refinement is essential for cross-patch consistency.

\subsubsection{Effect of Global Semantic Context in SARM}

We further investigate the role of global semantic context within the SARM. As shown in Table~\ref{table:abla_sarm}, without global semantics, SARM relies solely on self-attention for inter-patch feature exchange, which already improves consistency over the variant without SARM (Model2). Incorporating cross-attention with LR global semantics further boosts MUSIQ by 2.34 and reduces NIQE by 0.29, demonstrating that global context complements local inter-patch fusion for coherent reconstruction.

\begin{table}[!ht]
    \caption{Ablation on global semantic context within SARM. \label{table:abla_sarm}}
    \centering
    \scriptsize
    \resizebox{1\linewidth}{!}{
    \begin{tabular}{l |c c c c}
    \toprule
    \multirow{2}{*}{\textbf{Configuration}} & \multicolumn{4}{c}{\textbf{$\times$ 18} ($\times 4 \times 3 \times 1.5$)} \\
      {} & \bf{LPIPS$\downarrow$} & \bf{MUSIQ$\uparrow$} & \bf{NIQE$\downarrow$} & \bf{PI$\downarrow$} \\
    \midrule
     w/o global semantic &  0.469 & 49.10 & 6.30 & 5.49 \\
     w/ global semantic &  \textbf{0.450} & \textbf{51.44} & \textbf{6.01} & \textbf{5.24} \\
    \bottomrule
    \end{tabular}
    }
\end{table}

\subsubsection{Loss Function Analysis}

\begin{table}[t]
    \caption{Ablation study on loss functions, validating the effectiveness of each loss component.
    \label{table:loss}}
    \centering
    \scriptsize
    \resizebox{1\linewidth}{!}{
    \begin{tabular}{c c | c c c c} 
    \toprule
      \multirow{2}{*}{\textbf{$L_{corr}$}} & \multirow{2}{*}{\textbf{$L_{depth}$}}  & \multicolumn{4}{c}{\textbf{$\times$ 18} ($\times 4 \times 3 \times 1.5$)} \\ 
    {} & {} &\bf{LPIPS$\downarrow$} & \bf{MUSIQ$\uparrow$} & \bf{NIQE$\downarrow$} & \bf{PI$\downarrow$} \\ 
    \midrule
    {\color{blue}\ding{55}}  & {\color{blue}\ding{55}} & 0.462 & 49.33 & 6.71 & 5.91 \\
     
     {\color{red}\ding{51}}  & {\color{blue}\ding{55}}  & 0.459 & 50.23 & 6.24 & 5.40 \\

     {\color{red}\ding{51}}  & {\color{red}\ding{51}} &  0.450 & 51.44 & 6.01 & 5.24 \\
    \bottomrule
    \end{tabular}
    }
\end{table}

Table~\ref{table:loss} shows the contribution of each auxiliary loss.
$L_{\text{rec}}$ serves as the base reconstruction objective; removing both auxiliary terms already yields reasonable results.
Adding $L_{\text{corr}}$ enforces the self-similarity structure of the reconstructed image to match the ground truth, improving MUSIQ by 0.90 and NIQE by 0.47.
Further incorporating $L_{\text{depth}}$ provides geometric consistency, yielding the best overall performance.

\subsubsection{Superpixel Size Analysis}
\begin{figure}[t]
\centering
\begin{minipage}[h]{1\linewidth}
    \begin{minipage}[h]{1\linewidth}
        \centering
        \begin{minipage}[h]{0.185\linewidth}
            \centering
            \includegraphics[width=1\linewidth]{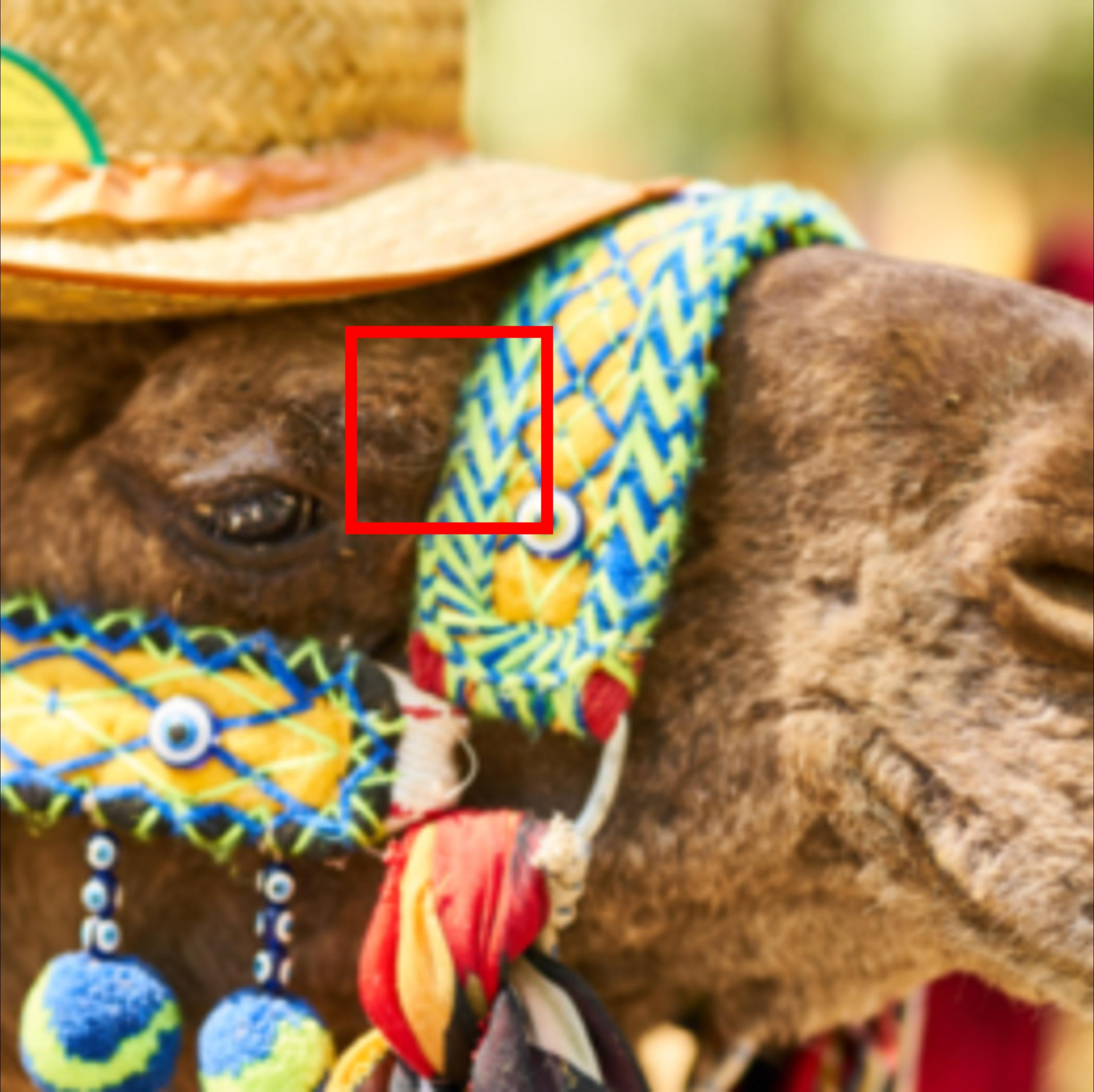}\\  
            \scriptsize{Origin}
        \end{minipage}
        \hfill
        \begin{minipage}[h]{0.185\linewidth}
            \centering
            \includegraphics[width=1\linewidth]{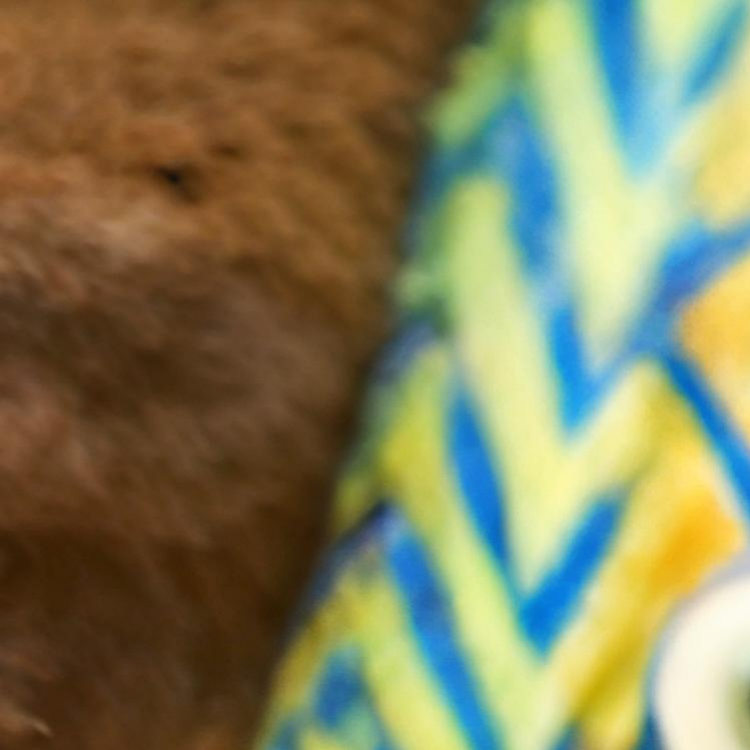}\\   
            \scriptsize{3 $\times$ 3}
        \end{minipage}
        \hfill
        \begin{minipage}[h]{0.185\linewidth}
            \centering
            \includegraphics[width=1\linewidth]{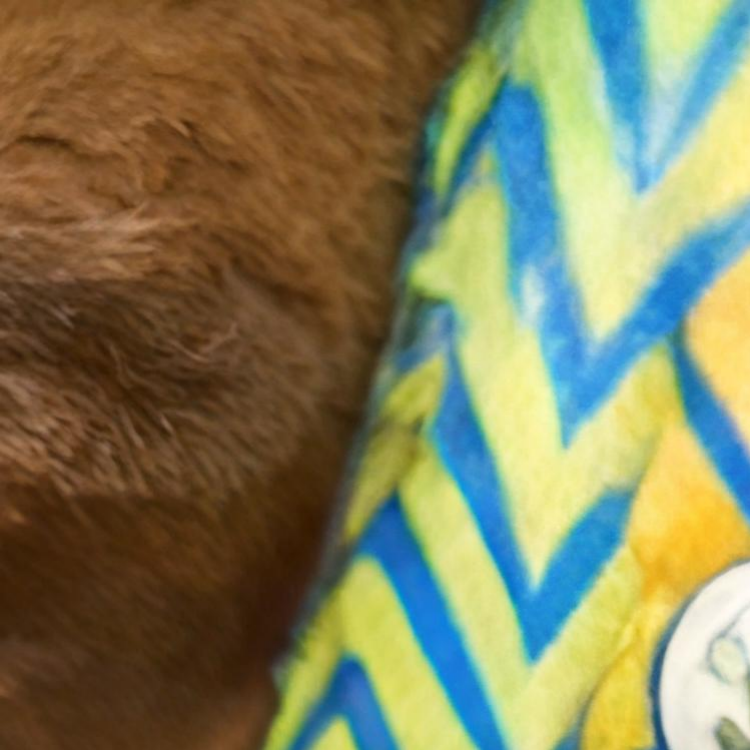}\\
            \scriptsize{4 $\times$ 4}
        \end{minipage}
        \hfill
        \begin{minipage}[h]{0.185\linewidth}
            \centering
            \includegraphics[width=1\linewidth]{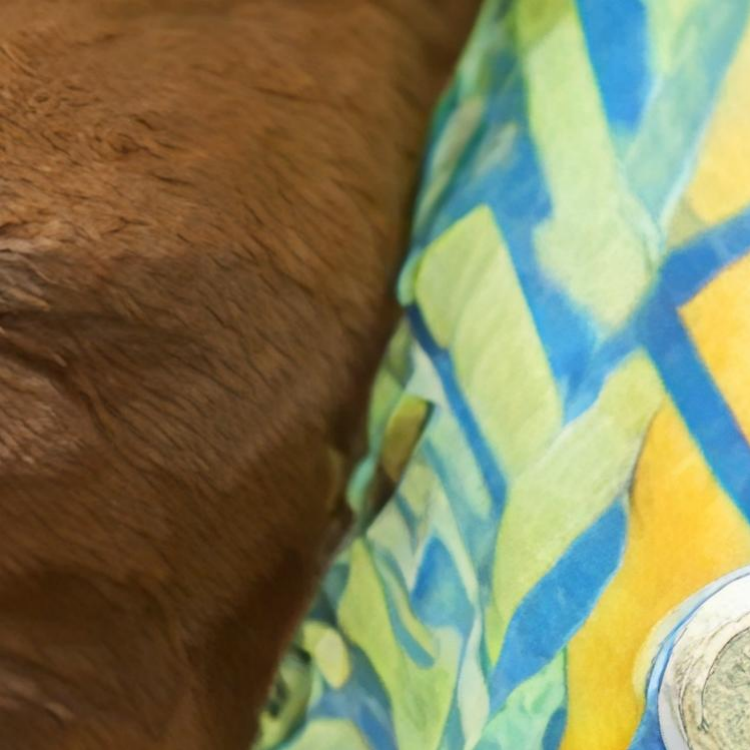}\\
            \scriptsize{5 $\times$ 5}
        \end{minipage}
        \hfill
        \begin{minipage}[h]{0.185\linewidth}
            \centering
            \includegraphics[width=1\linewidth]{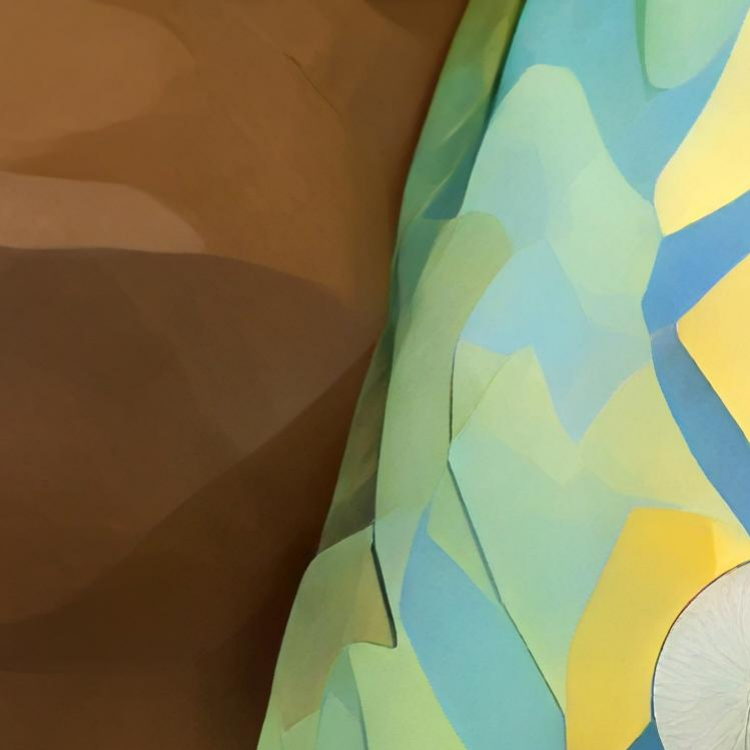}\\
            \scriptsize{8 $\times$ 8}
        \end{minipage}
    \end{minipage}
\end{minipage}

\centering
\caption{While superpixels effectively suppress degradation artifacts, excessively large superpixel sizes remove fine details and may even alter image content.}\label{fig:strokes}
\vspace{-6pt}
\end{figure}

Table~\ref{table:strokes} and Fig.~\ref{fig:strokes} summarize the effect of superpixel block size on performance. As the superpixel size increases, perceptual quality improves while structural consistency decreases, suggesting that larger superpixels more effectively suppress cascading artifacts but tend to remove fine details. We adopt a \(4 \times 4\) size for a balanced trade-off.

\begin{table}[t]
    \scriptsize
    \centering
    \caption{Ablation study on superpixel size. \label{table:strokes}} 
    \resizebox{1\linewidth}{!}{
    \begin{tabular}{  c |c  c c c }
        \toprule
        \multirow{2}{*}{\textbf{SuperPixel Size}} & \multicolumn{4}{c}{DIV8K}\\
         {}  & \textbf{LPIPS$\downarrow$} & \textbf{MUSIQ$\uparrow$} & \textbf{NIQE$\downarrow$}  &  \textbf{PI$\downarrow$}  \\
        
        \midrule
         3 $\times$ 3 &  0.513 & 35.11 & 7.67 & 6.81 \\
         4 $\times$ 4 &  0.450 & 51.44 & 6.01 & 5.24 \\
         5 $\times$ 5 &   0.481 & 53.19 & 5.33 & 4.42 \\
         8 $\times$ 8&  0.516 & 64.29 & 6.29 & 4.89 \\   
        \bottomrule

    \end{tabular}
    }
\end{table}


\section{Conclusion}
We demonstrate that ASISR becomes fundamentally more stable when ultra-magnification is modeled as a sequence of distribution-consistent transitions rather than a single extrapolation step. This shift reframes ASISR as a principled and inherently scalable paradigm, revealing that the key to extreme-resolution reconstruction lies not in enlarging models or datasets, but in understanding and regulating how representations evolve across scales.
Beyond its empirical benefits, this distribution-aware cyclic perspective may open directions for future research. It provides a conceptual foundation for unified multi-scale generative models, progressive detail synthesis, and controllable magnification, and may extend naturally to video, 3D content, and cross-modal reconstruction. Promising directions include adaptive superpixel sizing for content-aware structural alignment and extending the cyclic framework to other generative tasks.

\section*{Acknowledgments}
This work was supported in part by National Key R\&D Program of China 2023YFF0905103.\par

{
    \small
    \bibliographystyle{ieeenat_fullname}
    \bibliography{main}
}


\end{document}